\documentclass[journal]{IEEEtran}
\usepackage{times}
\usepackage{textcomp}
\usepackage{graphicx}
\usepackage{amsmath}
\usepackage{amssymb}
\usepackage{multirow}
\usepackage[caption=false]{subfig}

\newcommand{\E}{\mathbb{E}}
\DeclareMathOperator*{\argminA}{arg\,min}
\newcommand{\argmin}[1]{\underset{#1}{\argminA}\;}
\begin{document}

\title{Fast and Robust Cascade Model for Multiple Degradation Single Image Super-Resolution}

\author{Santiago L\'opez-Tapia, \and
Nicolás P\'erez de la Blanca\\
Dept. of Computer Science and Artificial Intelligence, University of Granada, Granada, Spain\\
{\tt\small sltapia@decsai.ugr.es, nicolas@ugr.es}}

\author{
Santiago L\'opez-Tapia, \and
Nicolás P\'erez de la Blanca
\thanks{Santiago López-Tapia and Nicolás P\'erez de la Blanca are with Dpto. de Ciencias de la Computaci\'on e I. A., Universidad de Granada, Spain.}
\thanks{This work was supported in part by the Spanish Ministerio de Econom\'{\i}a y Competitividad under contract. Santiago López-Tapia received financial support through the Spanish FPU program.}}

\maketitle
\begin{abstract}
Single Image Super-Resolution (SISR) is one of the low-level computer vision problems that has received increased attention in the last few years. Current approaches are primarily based on harnessing the power of deep learning models and optimization techniques to reverse the degradation model. Owing to its hardness, isotropic blurring or Gaussians with small anisotropic deformations have been  mainly considered. Here, we widen this scenario by including large non-Gaussian blurs that arise in real camera movements. Our approach leverages the degradation model and proposes a new formulation of the Convolutional Neural Network (CNN) cascade model, where each network sub-module is constrained to solve a specific degradation: deblurring or upsampling. A new densely connected CNN-architecture is proposed where the output of each sub-module is restricted using some external knowledge to focus it on its specific task. As far we know this use of domain-knowledge to module-level is a novelty in SISR. To fit the finest model, a final sub-module takes care of the residual errors propagated by the previous sub-modules. We check our model with three state of the art (SOTA) datasets in SISR and compare the results with the SOTA models. The results show that our model is the only one able to manage our wider set of deformations. Furthermore, our model overcomes all current SOTA methods for a standard set of deformations. In terms of computational load, our model also improves on the two closest competitors in terms of efficiency. Although the approach is non-blind and requires an estimation of the blur kernel, it shows robustness to blur kernel estimation errors, making it a good alternative to blind models.
\end{abstract}


\section{Introduction}
\label{sec:introduction}
Single Image Super-Resolution (SISR) is a fundamental low-level computer vision that has received considerable attention in recent years \cite{Yang2014SingleImageSA, dong2014learning, huang2015self, Kim2016, lim2017enhanced, Zhang2017learning, SFTGANCVPR18, zhang2019deep}. It consists of recovering a high-resolution (HR) image from a given low resolution (LR) image, assuming a degradation model connecting both images. Typically, the general image degradation model assumed in the SISR literature is given by, 
\begin{equation}\label{eq:im_for}
y = [x\circledast k]\downarrow_s + n,
\end{equation}
where $x$ is the HR image, $y$ is the LR image, $n$ is the noise, usually additive white Gaussian noise (AWGN), $\downarrow_s$ is the downsampling operator for factor $s$ and $x\circledast k$ represents the convolution of $x$ with the blur kernel $k$.

The current models used to estimate $x$ from $y$ are typically grouped into three categories according to the assumed degradation model and the solver used: interpolation-based (IB), model-based or energy-function optimization (MBO) and learning-based (LB). The IB approach assumes the following simpler model as the degradation model, 
\begin{equation}\label{eq:im_for_simple}
y= x\downarrow_s + n,
\end{equation}
and an interpolation technique is proposed as the solver. Clearly, Eq.~(\ref{eq:im_for_simple}) models the downsampling operation including as part of the noise any other degradation present in the image. Although these methods are the fastest, they are too simple to cope with real blur degradation effects \cite{zhang2018learning, zhang2019deep, gu2019blind}.

Traditional MBO approaches \cite{katsaggelos2007super, HE2009MAPSR, Tai2010, belekos2010maximum,babacan2011variational} define a regularized energy function that is optimized to estimate $x$. The degradation model (\ref{eq:im_for}), together with smoothness or edge preservation conditions in solution, are the most common elements that define the energy function \cite{babacan2011variational,Tai2010}. In these models, an optimization is performed for each new LR image, which provides them with high flexibility, under the LR imaging generation conditions, to manage any degradation type. However, closed-form solutions are not feasible and the use of iterative optimization techniques makes MBO approaches computationally expensive. In addition, the hyperparameters must be hand-picked for each degradation type \cite{Romano2017TheLE}.

In recent years, LB approaches have gained significant momentum thanks to the introduction of deep learning (DL). Recently, variable splitting techniques such as half-quadratic splitting (HQS) \cite{Geman1995} and alternating direction method of multipliers (ADMM) \cite{Boyd:2011} have been used to incorporate DL in MBO to improve efficiency. In \cite{Zhang2017learning}, \cite{Meinhardt2017LearningPO}, and \cite{bigdeli2017deep}, for instance, a denoising CNN was used to estimate the final HR image. Furthermore, in \cite{zhang2019deep}, the authors expand on this approximation by tasking the CNN with upsampling and denoising in a non-blind approach. Clearly, this introduces a connection between MBO and CNN-based approaches because the MBO solution eventually relies on learned modules. However, this connection assumes a new degradation model,
\begin{equation}
    \label{degradation_new}
     y= x\downarrow_s\circledast k^L + n,
\end{equation}
where blurring $k^L$ is applied on a downsampled image. Until now, no connection has been provided between this new model and the one in Eq.~(\ref{eq:im_for}).

In contrast to MBO approaches, most LB models do not explicitly use any image degradation model to estimate $x$, assuming that the degradation model is well represented by Eq.~(\ref{eq:im_for_simple}). A large database of pairs (LR, HR) is used to learn an end-to-end mapping $f(\cdot)$ such that $x = f(y)$ \cite{yang2012sr, zhang2013multiscale, trinh2014example, hui2015bimodal}. The DL models based on Convolutional Neural Networks (CNNs) have shown the highest performance compared to any other approach \cite{dong2014learning, Kim2016, lim2017enhanced, SFTGANCVPR18, Zhang2019Coll}. The early success of vanilla CNN models in SISR \cite{dong2014learning,dong2016fast} has stimulated the development of deeper designs that have achieved the current state of the art (SOTA) in SISR. As a matter of fact, most of the proposals from the first one, given in \cite{dong2014learning}, have been driven by the incorporation of technical achievements in the CNN-field as for instance, deeper models, and use of residual block to improve accuracy metrics \cite{deepSurvey2020,kim2016accurate, ledig2016photo, lim2017enhanced, SFTGANCVPR18, zhang2018DSR, zhang2018rcan, Wang2018ESRGANES}. To improve the performance and processing time, upsampling moves to the last layer of the network \cite{dong2016fast, Shi2016RealTimeSI,ledig2016photo}. In addition, SRGAN \cite{ledig2016photo} and EDSR \cite{lim2017enhanced} improved over those architectures and introduced residual blocks to increase the network depth. DenseSR \cite{zhang2018DSR} used residual dense blocks to improve the networks by combining features from different layers. In ESRGAN \cite{Wang2018ESRGANES} and RCAN \cite{zhang2018rcan} specific blocks to improve performance are proposed, reaching a new SOTA, but at the cost of a significant increase in computing time. Although these models outperform traditional MBO in terms of performance, they lack adaptation to new deformations at test time.

Very recently, CNN-based models (CNN-BMs) assuming Eq.~(\ref{eq:im_for}) as a degradation model have been proposed. In this scenario, Shocher et al. \cite{ZSSR} proposed a technique for the ``Zero-Shot'' case that exploits the internal recurrence of information inside a single image to train a DL model. This shares the flexibility at test time, but also the high time consumption of MBO approaches. Alternatively, recent works have proposed architectures that encode information about the blur kernel $k$ and introduce it into the network. Riegler et al. \cite{Riegler2015} used conditioned regression models to encode a Gaussian family of kernels. In SRMD \cite{zhang2018learning}, a Principal Component Analysis (PCA) representation of the blur kernel and a stretching strategy to concatenate it with the LR image is proposed. In \cite{gu2019blind}, the latter approach was expanded and improved, with the use of spatial feature transform (SFT) layers \cite{SFTGANCVPR18}, to introduce the PCA representation of the blur kernel into the network. By doing so, they are able to implement a very deep residual network with a significant increase in performance over \cite{zhang2018learning}. However, PCA is not a suitable representation to encode families of blurs with different degrees of freedom, in addition to having to be recalculated to cope with new degradation types. Consequently, the CNN models used in \cite{zhang2018learning} and \cite{gu2019blind} show shortcomings when blur kernels from strong camera movement are present \cite{zhang2018learning}.

An alternative to the above CNN-BMs that attempt to solve the SISR problem using a single CNN is to decompose the problem into sub-tasks and use a group of CNNs connected in a cascade fashion to solve it. This ``divide and conquer'' approach simplifies the function that each sub-module must learn, easing the learning process. If $n=0$, Eq.~(\ref{eq:im_for}) can be inverted using an upsampling CNN followed by a deblurring CNN. However, in practice, this approximation does not perform well because of the introduction of strong artifacts during upsampling and the increasing difficulty of deconvolving in HR space \cite{gu2019blind}. The CNN-BMs based on
cascades \cite{gu2019blind}, \cite{zhang2019deep}, \cite{Yuan2018UnsupervisedIS} make the first denoise/deblur and later upsampling implicitly assuming that Eq.~(\ref{eq:im_for}) can be approximated by Eq.~(\ref{degradation_new}). In \cite{Yuan2018UnsupervisedIS}, an unsupervised network was trained to deblur and denoise the LR image, before using an SR network to estimate the HR. However, as shown in \cite{gu2019blind} and \cite{zhang2019deep}, these cascade approaches underperform compared with other SOTA models. Here, we argue that cascade models can be significantly improved by modifying the connection between the sub-modules and taking advantage of the fact that each sub-module can be specialized in a task by constraining its output using domain knowledge.

In this study, we propose a fast, accurate, and robust CNN cascade-based approach for multiple degradations SISR, CAscade-Deblurring-Upsampling-Fusion (CADUF). Compared with previous cascade-based approaches, our model shows new elements regarding deblur, upsampling, and information processing. In deblurring, the LR image is complemented with the information provided by a deblurred and noisy release of itself obtained from the Weiner filter. We take advantage of the fact that artifacts and noise introduced by the Weiner filter are similar across all possible combinations of blur and images. After extracting and matching features from the two images, we use those features to filter out the noise and artifacts in the Weiner-filtered image.
In upsampling, we make use of domain knowledge in the form of ``Privileged Information'' \cite{vapnik2015privi} and constrain the
output of upsampling to fulfill an affine projection model. As a result, we obtain an error range that can be reduced by a simple final sub-module. Finally, we improve the communication between sub-modules by using a novel connection approach where each sub-module re-uses the features calculated by all previous sub-modules. Our experiments for non-blind SISR show that CADUF reaches SOTA results being competitive with MBO for unseen blurs but much faster. Furthermore, experiments performed in the blind setting show that our method is also robust to inaccuracies in the blur kernel estimation.

The remainder of the paper is organized as follows. In Section~\ref{sec:model_description}, we describe the proposed CADUF model and explain each one of its primary components. Section~\ref{sec:experimental_results} presents the experimental settings used in this study (dataset, degradations and training hyperparameters) as well as an ablation study of each key component and a comparison with SOTA methods. Finally, Sect.~\ref{sec:conclusions_results} concludes the paper.

\section{Method}
\label{sec:model_description}
In this section, we first introduce the SR problem settings and propose a new CNN architecture that addresses it by combining multiple CNN sub-modules that solve specific sub-problems. Each of these CNN sub-modules are presented in Sections ~\ref{sec:motion_and_features} to \ref{sec:refinement}. Finally, in Section~\ref{sec:architecture_detail}, we describe in detail the parameters of the proposed architecture.

\begin{figure}[!htb]
\begin{center}
\setlength{\tabcolsep}{2pt}
\renewcommand{\arraystretch}{1}
\begin{tabular}{ccccccc}
\includegraphics[width=0.06\textwidth]{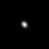} &
\includegraphics[width=0.06\textwidth]{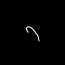} &
\includegraphics[width=0.06\textwidth]{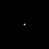} &
\includegraphics[width=0.06\textwidth]{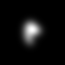} &
\includegraphics[width=0.06\textwidth]{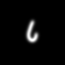} &
\includegraphics[width=0.06\textwidth]{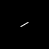} &
\includegraphics[width=0.06\textwidth]{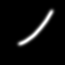} \\
\end{tabular}
\end{center}
\caption{Examples of the blur kernels $k$ considered in this work. They represent a blur combination of isotropic Gaussian and motion.}
\label{fig:example_kernels}
\end{figure}

As previously stated in Section~\ref{sec:introduction}, we consider that the LR image $y$ is produced from the high-res image $x$ in Eq.~(\ref{eq:im_for}). Typically, in SISR, $k$ is assumed to be an isotropic Gaussian kernel and $\downarrow_s$ bicubic interpolation \cite{zhang2018learning, gu2019blind}, but in real applications, different kernel types could arise. In this study, we consider that $k^H$ can also be a more complex blurs, like motion blur. Fig.~\ref{fig:example_kernels} shows some examples of the complexity of the kernels used during our experiments.

We propose a CNN architecture composed of several CNN sub-modules, each of which focuses on solving a portion of the degradation process that produces the LR image. Using this approach has several advantages over using a current SR CNN:

\begin{enumerate}
\item By breaking the SR task into sub-tasks each addressed by a sub-module, a strong regularization is imposed over the function space that the model is able to explore. This makes learning easier, leading to improved optima and generalization.
\item Task-specific constraints in each sub-module can be introduced, to further reduce the search space (see Sections~\ref{sec:deblur} and \ref{sec:sr}).
\item Specialized architectural elements can be used for each task in each sub-module, improving model performance and efficiency. (see Section~\ref{sec:motion_and_features}).
\end{enumerate}

To do this, we design a cascade based on Eq (\ref{degradation_new}) as an approximation to Eq (\ref{eq:im_for}). This allows us to first perform denoising and deblurring before we upsample the LR image. Based on Eq.~(\ref{degradation_new}) and assuming that $k^L$ is known, we propose the decomposition of the architecture in the following specialized sub-modules:
\begin{enumerate}
\item $\mathrm{E}_\theta$: Extraction of motion-corrected features. See Section~\ref{sec:motion_and_features}.
\item $\mathrm{D}_\phi$: Deblur and Denoising of the LR image, Section~\ref{sec:deblur}.
\item $\mathrm{U}_\psi$: Upsampling, to get an initial SR solution, Section~\ref{sec:sr}.
\item $\mathrm{F}_\omega$: Refinement, improvement of the initial SR solution by cleaning the errors and artifacts caused by the accumulated errors in the cascade. See Section~\ref{sec:refinement}.
\end{enumerate}

\begin{figure}[!htb]
    \begin{center}
    \includegraphics[width=0.45\textwidth]{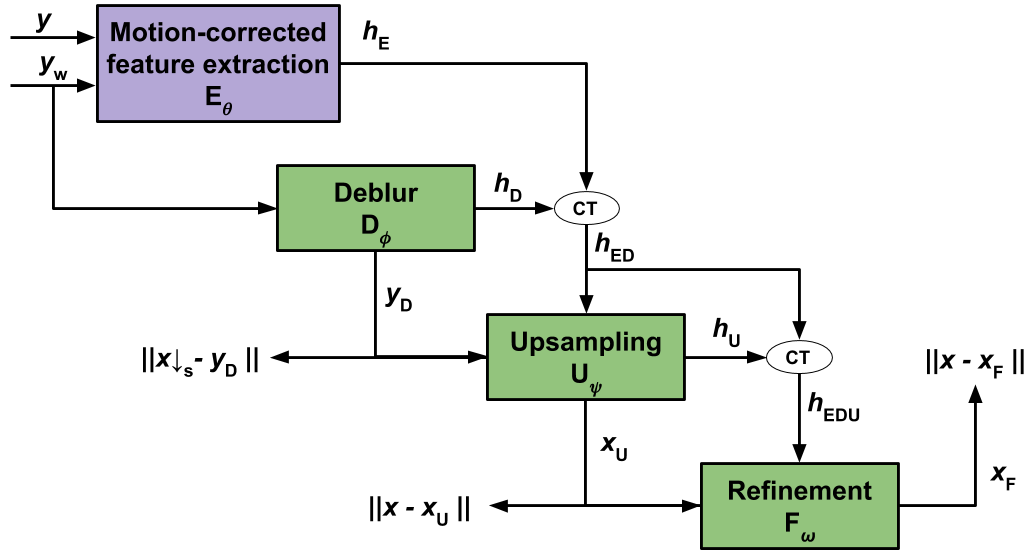}
    \end{center}
    \caption{Proposed architecture for solving the MDSR problem. Each sub-module is a CNN and is specialized in solving a sub-task. Note that ``CT'' indicates concatenation of the feature maps.}
    \label{fig:fusion_models}
\end{figure}

The structure of the proposed architecture is shown in Fig.\ref{fig:fusion_models}. The architecture follows a cascade approach where the features $h$ and the result of each sub-task are needed for the next sub-module in the cascade. Compared to other cascade approaches, our sub-modules use not only the previously calculated image, but also the features calculated by all previous sub-modules, which allows the propagation and reuse of more information through the cascade. For each spatial location, our model encodes a vector of information, instead of just a value, better representing the distribution of solutions. The architecture is trained in a end-to-end fashion, being its loss function
\begin{equation}
\label{eq:loss}
{\cal L}_\mathrm{CADUF} = \alpha{\cal L}_\mathrm{D} + \beta{\cal L}_\mathrm{U} + (1-\alpha-\beta){\cal L}_\mathrm{F}
\end{equation}
where $\alpha,\beta>0$, $\alpha+\beta<1$ and ${\cal L}_\mathrm{D}$, ${\cal L}_\mathrm{U}$ and ${\cal L}_\mathrm{F}$ are specific losses for sub-modules $\mathrm{D}_\phi$, $\mathrm{U}_\psi$ and $\mathrm{F}_\omega$, respectively. These specific losses will be defined later.

Let us now describe in detail each of the proposed sub-modules of the architecture.

\subsection{Extraction of motion-corrected features}
\label{sec:motion_and_features}
Following Eq.~(\ref{degradation_new}), the first issue to be addressed is the deconvolution of $y$ by the blur kernel $k^L$ to obtain $x\downarrow_s$. Because the deformations present in the image depend on the combination of the blur kernel $k^L$ and the image, learning a single CNN that can cope with all these variations is a challenging task. For this reason, we propose to start from an initial noisy solution that can be calculated quickly using the Wiener filter:
\begin{equation}
    y_w = {\cal F}^{-1}(
    \frac{\overline{{\cal F}(k^L)}{\cal F}(y)}
    {\overline{{\cal F}(k^L)}{\cal F}(k^L) + \epsilon}
    ),
\end{equation}
where ${\cal F}$ and ${\cal F}^{-1}$ denote the fast Fourier transform (FFT) and inverse FFT, respectively, and $\overline{{\cal F}}$ denotes the complex conjugate of ${\cal F}$, and $\epsilon$ is the regularization term.

\begin{figure}[!htb]
\captionsetup[subfigure]{labelformat=empty}
\begin{center}
\setlength{\tabcolsep}{1pt}
\renewcommand{\arraystretch}{0.5}
\begin{tabular}{cccc}
\multirow{2}{*}[5.2em]{\subfloat[Sampled GT ($x\downarrow_s$)]{\includegraphics[width=0.196\textwidth]{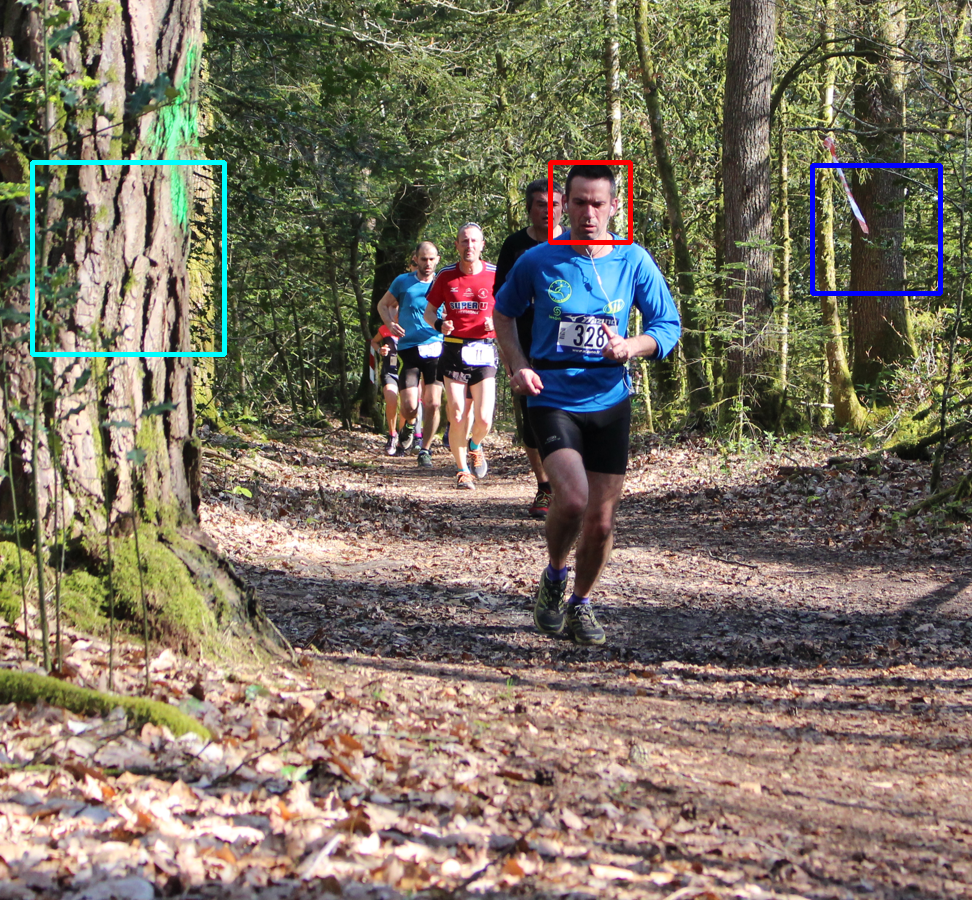}}} &
\includegraphics[width=0.09\textwidth]{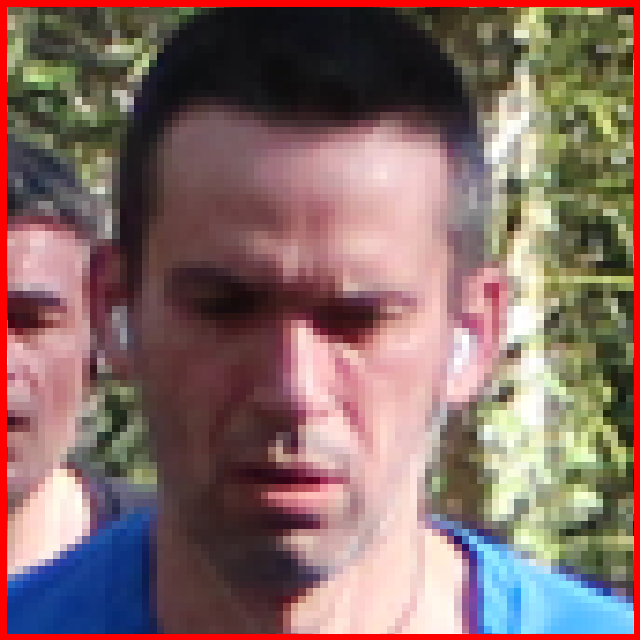} &
\includegraphics[width=0.09\textwidth]{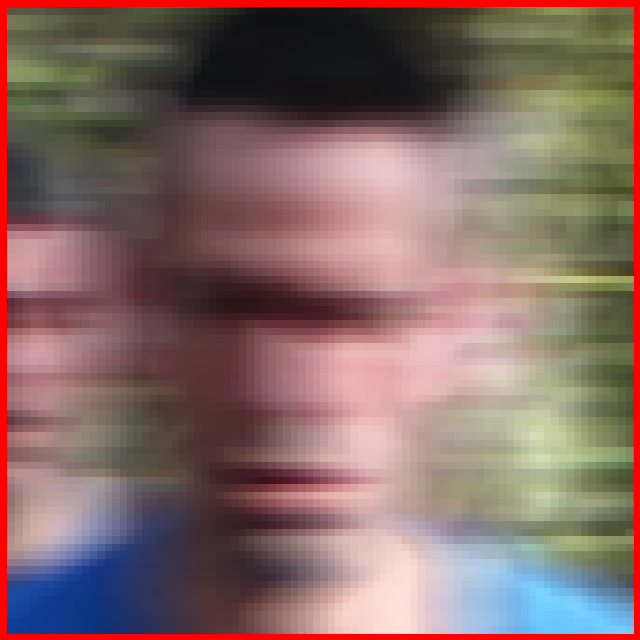} &
\includegraphics[width=0.09\textwidth]{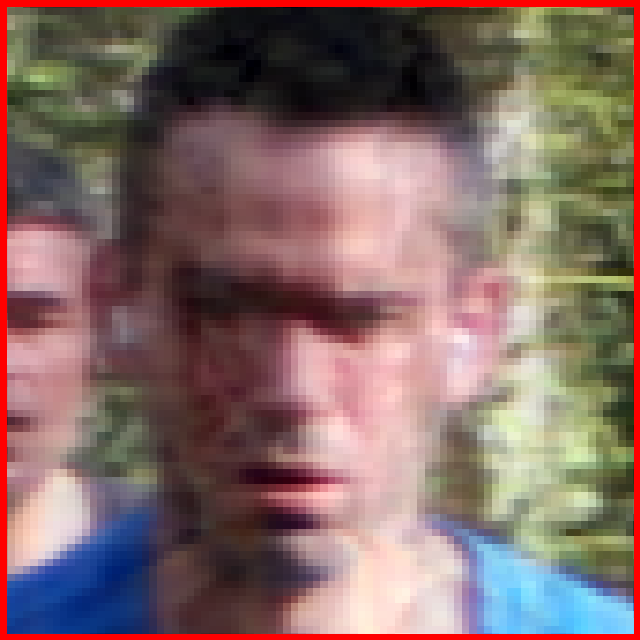} \\ [-0.2ex] &
\subfloat[Sampled GT]{\includegraphics[width=0.09\textwidth]{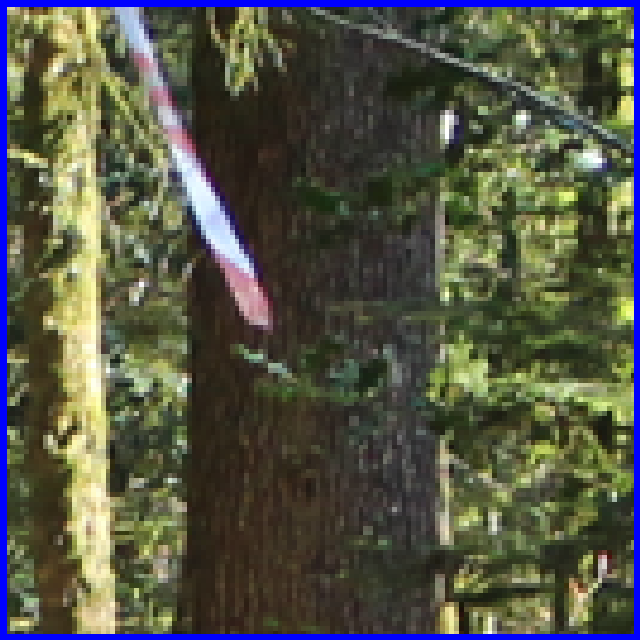}} &
\subfloat[LR ($y$)]{\includegraphics[width=0.09\textwidth]{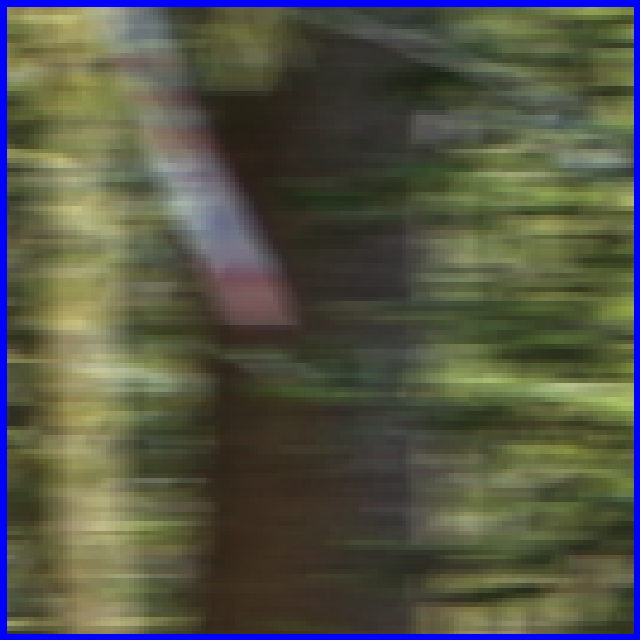}} &
\subfloat[Wiener ($y_w$)]{\includegraphics[width=0.09\textwidth]{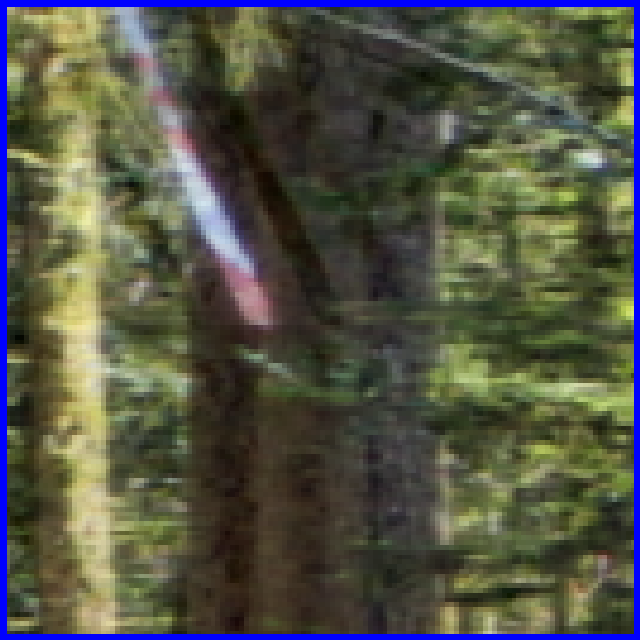}} \\ [-2ex]
\end{tabular}
\end{center}
\caption{Illustration of the artifacts originating from using the Wiener filter. Some produced artifacts can be compatible with natural images, being difficult to distinguish from real structures using only the filtered image.}
\label{fig:wiener_artifacts}
\end{figure}

As seen in Section~\ref{sec:ablation}, the use of $y_w$ as the starting point significantly outperforms using only the blur image $y$. However, the Wiener filter increases the noise in the image and introduces artifacts that are difficult to eliminate without the information of the blur image $y$, see Fig~\ref{fig:wiener_artifacts}. To correct the image with a CNN, at each spatial location, we use the features extracted from $y$ and $y_w$. However, because of the blurring processes, $y$ and $y_w$ are not aligned with each other. In this case, we can consider $y_w$ as an anchor that indicates the correct position.

To align the features extracted from both $y$ and $y_w$, we propose the use of deformable convolution. First proposed in \cite{dai17dcn} and enhanced in \cite{zhu18dcn} to handle more deformations, deformable convolutions are a modified convolution operation:
\begin{equation}\label{eq:dcnn}
    g^l(m) = \sum_{o=1}^O w_{o}a_{m,o}f^{l-1}(m+o+\Delta m_{m,o}),
\end{equation}
where $g^l(m)$ denotes the feature vector in layer $l$ at location $m$, $o$ is the position of the convolutional kernel $w$, and $\Delta m_{m,o}$ and $a_{m,o}$ are offsets and scalar modulation, respectively. These parameters are calculated for each $m$ location using another convolutional layer. Owing to this operation, our model can align the features of $y$ and $y_w$ at each spatial location, making further downstream calculations much easier.

\begin{figure}[!htb]
    \begin{center}
    \includegraphics[width=0.45\textwidth]{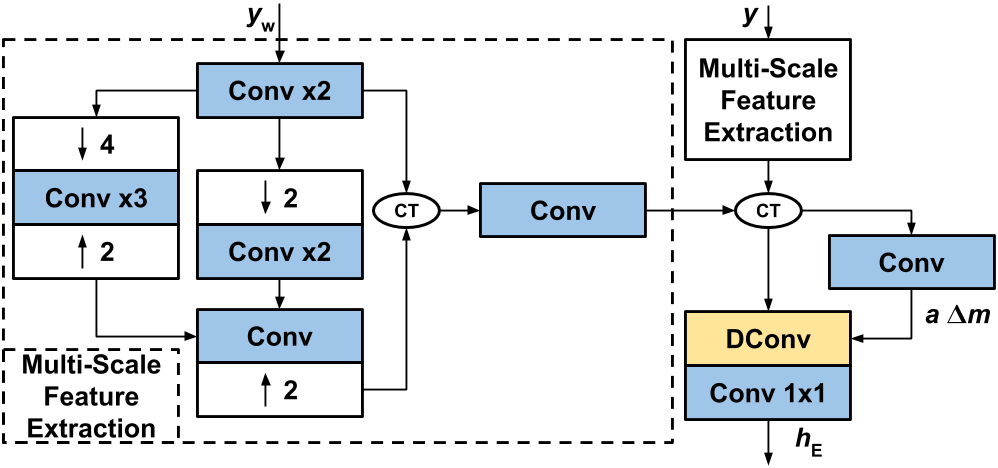}
    \end{center}
    \caption{Motion-corrected feature extraction sub-module. Conv is a convolution, and DConv is a deformable convolution. Each convolution is followed by a leaky-ReLU activation with $0.2$ of negative slope and has a $3\times3$ kernel with the exception of the first and last convolutions that use a $5\times5$ and $1\times1$, respectively. Convolutions for scale $1$ use 64 features while convolutions on scales $/2$ and $/4$ use 32.}
    \label{fig:motion_correction}
\end{figure}

Nevertheless, the calculation of parameters $a$ and $\Delta m$ is not an easy task: Movement of the camera can be quite strong, thus matching locations between $y$ and $y_w$ can be too far away. In these cases, a large receptive field is required. To obtain this, we extract features at scales $1$, $/2$, and $/4$ using the architecture shown in Fig.~\ref{fig:motion_correction}. By doing so, we increase the receptive field to $37\time37$ pixels without the need to use a far deeper architecture.

This sub-module, $\mathrm{E}_\theta$, produces the features $h_{\mathrm{E}}$ that will be used by the remaining network. Although it could move the feature vectors anywhere, by propagating the loss of the next sub-modules in the cascade, it would have to learn to put then in the correct place, producing motion-corrected features.

\subsection{LR-Anchor image estimation}
\label{sec:deblur}
According to Eq.~(\ref{degradation_new}), we start mapping the LR image $y$ to an LR image $y_\mathrm{D}^*$ without blur or noise.

To define $y_\mathrm{D}^*$, one approach can be defining it as $x\downarrow_s$. However, the difficulty of the SR task primarily depends on the blur kernel used before downsampling. Those cases with both low and high Gaussian blur led to significantly worse reconstruction. Large kernels in HR produce highly blurry LR images while small kernels in HR do not have a smooth downsampling form in LR, which introduces aliasing artifacts in the LR image. Aliased LR images can be very difficult to super-resolve because artifacts must be distinguished from real high-frequency patterns in the image.

Following the above arguments, we define $y_\mathrm{D}^*$ as the LR image obtained by downsampling $x$ with a downsampling operator $\downarrow_s^\mathrm{D}$ that it is easier to invert by the upsampling network that we use. That is:
\begin{equation}
y_\mathrm{D}^*= x\downarrow_s^\mathrm{D} = [x\circledast k^\mathrm{D}]\downarrow_s,
\end{equation}
where $\downarrow_s$ is the bicubic downsampling of factor $s$, and $k^\mathrm{D}$ is an isotropic Gaussian with $\sigma$ $0.8$ and $1.8$ for scaling factors 2 and 4, respectively. These values have been found experimentally from within the range [$0.2$, $4.0$].
Notice that the selection of these values can be seen as a form of ``Privileged Information'' \cite{vapnik2015privi}. The images $\{y_\mathrm{D}^*\}$ represent additional information available only during training and are used to estimate the optimal LR image. In this regard, $y_\mathrm{D}^*$ acts as an anchor that guides and regularizes the model training.

The features for the deblurring task are computed using a CNN-regression model that maps its input $\{h_{\mathrm{E}}, y_w\}$, the motion-corrected features, and the low-resolution Wiener deconvolved image, into $y_\mathrm{D}^*$, the downsampled version of $x$. Let us denote by $\mathrm{D}_\phi(h_{\mathrm{E}}, y_w)$ the mapping defined by the CNN model. Then, $\mathrm{D}_{\hat{\phi}}(h_{\mathrm{E}}, y_w):\{y_w\}\rightarrow \{y_\mathrm{D}\}$, where $\hat{\phi}$ represents the weights learned after training the model.
The parameters can be optimized introducing the following loss ${\cal L}_\mathrm{D}$
\begin{equation}
    \label{map-deconv}
     {\cal L}_\mathrm{D}=\frac{1}{N}\sum_{i=1}^N{\cal L}(\mathrm{D}_\phi(h_{\mathrm{E}}, y_w), y_\mathrm{D}^*),
\end{equation}
where ${\cal L}(\cdot)$ is any loss function that estimates the differences between two images, like mean squared error (MSE). It can be observed $\mathrm{D}_\phi$ computes a deblurring and denoising mapping for the degradation model defined in Eq.~(\ref{degradation_new}) because the low-resolution image $y$ is mapped into an image without blur or noise, $y_\mathrm{D}$, that is easier to super-resolve.

\subsection{Initial high-res image estimation}
\label{sec:sr}
After deconvolution, we are left with upsampling. This is equivalent to the problem defined in Eq.~\ref{eq:im_for_simple} when $n=0$ and $\downarrow_s^\mathrm{D}=\downarrow_s$. Let us define the minimization of the upsampling process as follows: 
\begin{equation}
\label{eq:map-sr-ideal}
    \hat{\psi}=\arg\min_{\psi}\frac{1}{N}\sum_{i=1}^N {\cal L}(\mathrm{P}_\psi(z), x),
\end{equation}
where $\mathrm{P}_\psi$ is a CNN with parameters $\psi$, and $z$ is the input that will be defined later. It should be noted that this approach is similar to the one used in other cascade methods \cite{gu2019blind,zhang2019deep}. If the error introduced by the previous steps in the cascade is low, it is expected that it will not significantly affect this step.
In this case, the simplest solution, therefore, would be to use as an upsampling sub-module a classic CNN for SISR such as SRResNet \cite{ledig2016photo} to minimize Eq.~\ref{eq:map-sr-ideal}. However, as shown by our experiments in Section~\ref{sec:ablation}, this is not the best solution in practice. By minimizing Eq.~\ref{eq:map-sr-ideal}, it has to solve two distinctive tasks: Inversing the operator $\downarrow_s^\mathrm{D}$ and removing the approximation error. The learning of the parameters is more difficult: Small differences can be due to approximation errors or different structures in the HR space. If not constrained, small changes in the LR image can lead to very different SR images, leading to an ill-conditioned model. This could result in the appearance of strong artifacts.

To use Eq.~\ref{eq:map-sr-ideal} to learn only the inverse of $\downarrow_s^\mathrm{D}$, we constrained the predicted image $x_\mathrm{U}$ to those solutions that satisfy,
\begin{equation}
\label{eq:constraint}
x_\mathrm{U}\downarrow_s^\mathrm{D} = y_\mathrm{D},
\end{equation}
through the use of the projection introduced in \cite{sonderby2016amortised}. Let us define $A$ as a degradation operator such that $Ax=x\downarrow_s^\mathrm{D}$ and $A^+$ as the Moore-Penrose pseudoinverse \cite{Albert:book} of $A$ (refer to Section~\ref{sec:amas_calculation} to see the estimation of $A^+$). In our case, $A$ is an affine projection implemented as a stride convolution, and $A^+$ is implemented using a convolutional transpose layer. For a given $\mathrm{P}_\psi(z)$, a new transformation $\mathrm{U}_\psi(z)$ can be defined as,
\begin{equation}
\mathrm{U}_\psi(z) = (I-A^+A)\mathrm{P}_\psi(z) + A^+y_\mathrm{D},
\end{equation}
such that if no noise is present because $AA^+A=A$ and $A$ is a full row rank matrix, that is, $AA^+=I$, then,
\begin{equation}
A\mathrm{U}_\psi(z) = A(I-A^+A)\mathrm{P}_\psi(z) + AA^+y_\mathrm{D} = 0 + y_\mathrm{D}.
\end{equation}
Therefore, if $x_\mathrm{U}=\mathrm{U}_\psi(z)$, Eq.~\ref{eq:constraint} is always satisfied.

Owing to the loss of information caused by approximation errors, the information of $y_{\mathrm{D}}$ contained in $h_{\mathrm{D}}$ is not sufficient to properly invert $\downarrow_s^\mathrm{D}$. Our model needs information of the original $y$, which could not be preserved in $h_{\mathrm{D}}$. We add this information by concatenating to $h_{\mathrm{D}}$ the motion-corrected features $h_{\mathrm{E}}$, obtaining $h_{\mathrm{ED}}$. Note that this would not be possible without motion-correction because the features $h_{\mathrm{D}}$ will not align with features extracted from the blurred image $y$. Therefore, we define the input as $z=\{h_{\mathrm{ED}}, y_\mathrm{D}\}$. Using these inputs and the affine projection, we obtain a specialized SR sub-module by replacing $\mathrm{P}_\psi$ for $\mathrm{U}_\psi$ and introduced the following loss to optimized its parameters:
\begin{equation}
    \label{map-sr}
     {\cal L}_\mathrm{U}=\frac{1}{N}\sum_{i=1}^N{\cal L}(\mathrm{U}_\psi(h_{\mathrm{ED}}, y_\mathrm{D}), x).
\end{equation}

\subsection{High-res image refinement}
\label{sec:refinement}
Compared to previous SISR cascade models, ours adds an improvement in the form of a last fusion block fed with the characteristics of all the blocks in the cascade (see Fig.~\ref{fig:fusion_models}) focused on removing the accumulated error. This last block is the result of distinguishing two tasks in the previous step: Inverting the downsampling operator $\downarrow_s^\mathrm{D}$, in addition to removing the accumulated error $e$ introduced by the previous cascade sub-modules,
\begin{equation}
    \label{eq:approximation_error}
    x = x_\mathrm{U}+e.
\end{equation}
Note that $e$ can be seen as being structured noise dependent not only on $y$, but also on previous sub-modules $\mathrm{E}_\theta$, $\mathrm{D}_\phi$, and $\mathrm{U}_\psi$.
Because $\mathrm{U}_\psi$ is constrained, $e$ will not introduce strong artifacts when the error introduced by $\mathrm{D}_\phi$ is small (as shown in Section~\ref{sec:ablation}). Therefore, we argue that a last refinement sub-module $\mathrm{F}_\omega$, when fed with the features $h_\mathrm{EDU}$ calculated by each one of the specialized sub-modules, $\mathrm{E}_\theta$, $\mathrm{D}_\phi$, and $\mathrm{U}_\psi$, can learn to calculate a residual that, when added to $x_\mathrm{U}$, compensates for the accumulated errors $e$ in the cascade. We implemented $\mathrm{F}_\omega$ using a CNN regression model. The learning of the parameters is performed using the following loss function:
\begin{equation}
    \label{map-ref}
     {\cal L}_\mathrm{F}=\frac{1}{N}\sum_{i=1}^N{\cal L}(\mathrm{F}_\omega(h_{\mathrm{EDU}}, x_\mathrm{U}), x).
\end{equation}

\subsection{$\mathrm{D}_\phi$, $\mathrm{U}_\psi$ and $\mathrm{F}_\omega$ architecture}
\label{sec:architecture_detail}
\begin{figure}[!htb]
    \begin{center}
    \includegraphics[width=0.35\textwidth]{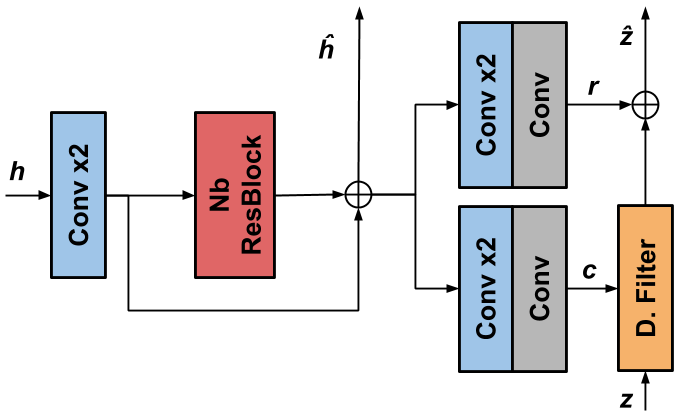}
    \end{center}
    \caption{Architecture used for sub-modules $\mathrm{D}_\phi$, $\mathrm{U}_\psi$, and $\mathrm{F}_\omega$. Conv indicates a convolutional layer and D. Filter the correlation of kernels $c$ with an input image $z$. We use $\mathrm{Nb}$ residual blocks \cite{PerSajHirSch18}, each composed of two convolutions after removing the batch normalization layers, as suggested in \cite{lim2017enhanced}. Each convolution uses a $3\times3$ kernel size, with the exception of the first layer that uses $1\times1$ kernels to fuse the previous sub-module features. With the exception of the output convolutions, all are followed by a leaky ReLU activation with a negative slope of $0.2$. We skip the first two convolutions for $\mathrm{D}_\phi$ because $\mathrm{E}_\theta$ already provides a suitable set of features. In the case of $\mathrm{U}_\psi$ and $\mathrm{F}_\omega$, sub-pixel convolutions \cite{Shi2016RealTimeSI} of scaling factor $s$ are used to produce $c$ and $r$ with the correct output size (see the text for further details).}
    \label{fig:architecture_detail}
\end{figure}

The architecture that implements our sub-modules $\mathrm{D}_\phi$, $\mathrm{U}_\psi$, and $\mathrm{F}_\omega$ is shown in Fig.~\ref{fig:architecture_detail}. As seen, it is a modification of the residual architecture SRResNet proposed in \cite{PerSajHirSch18}. We remove the batch normalization layers as suggested in \cite{lim2017enhanced} and replace the ReLU activation with leaky ReLU with $0.2$ negative slope. To further increase the receptive field, the first convolution in each residual block is a dilated convolution \cite{Yu:2016:MCA} with the dilatation parameter set to 2. The architecture is fed with features $h$ calculated by the previous sub-modules. The first two convolutional layers transform the input features. Then, $\mathrm{Nb}$ residual blocks are used to produce the new set of features $\hat{h}$. To calculate the output image $\hat{z}$, we use $\hat{h}$ together with a dynamic filter network (DFN) \cite{Xu2016DFN}. The DFN predicts $c$, which is a collection of filters, one for each spatial location in the output image. Using the approximation proposed in \cite{jo2018duf}, $\hat{z}$ is calculated as,

\begin{equation}
    \hat{z}_{i,j}
    = r_{i,j}+\sum_{l=-p}^{p}\sum_{m=-p}^{p}{c_{i,j,l,m}*z_{i+l,j+m}},
\end{equation}
where $z$ is the output image calculated by the previous sub-module, ${i,j}$ is a spatial location, and $r$ is a residual image calculated by a CNN using $\hat{h}$. We use a value of $p=2$. By using this approximation, our model can filter spatially-variant artifacts present in the outputs of previous sub-modules. Then, the filtered outputs can be used to perform residual learning \cite{zhang2017beyond, li2017video} more effectively. These artifacts are caused by approximation errors in the image formation model and the calculation. Notice that a normal CNN can also learn to remove these artifacts, but it would require far more parameters because CNN filters are spatially-invariant.

In each convolution, except in the input convolutions of each sub-module and the two output convolutions, kernels of size $3\times3 $ and 64 filters are used. The input convolutions use kernels of size $1\times1$ to fuse the features of the previous sub-modules. In our experiments, the number of residual blocks $\mathrm{Nb}$ was set to eight for $\mathrm{D}_\phi$, 16 for $\mathrm{U}_\psi$, and $\mathrm{F}_\omega$ uses only four. In the case of $\mathrm{U}_\psi$ and $\mathrm{F}_\omega$, both $c$ and $r$ are upsampled by scaling factor $s$ using sub-pixel convolutions \cite{Shi2016RealTimeSI}. The sub-pixel convolution is located after the last convolution for $c$ and before in the case of $r$. The first two convolutions of $\mathrm{D}_\phi$ are skipped because $\mathrm{E}_\theta$ already provides a suitable set of features.

\section{Experiments}
\label{sec:experimental_results}

\subsection{$k^{*L}$ estimation}
\label{sec:kl_calculation}
To train our models, for each $k$ in Eq.~(\ref{eq:im_for}), we need to have $k^{*L}$ for Eq.~(\ref{degradation_new}) such as $k^{*L} = \arg\min_{k^L} ||[x\circledast k]\downarrow_s - x\downarrow_s\circledast k^L||^2_F,$, that is, $k^{*L}$ minimizes the difference between both degradation models. To calculate $k^{*L}$, we use a neural network with one hidden layer with $2048$ neurons. This network $\mathrm{L}_\zeta$ with parameters $\zeta$ is trained by minimizing $||x\circledast k]\downarrow_s - x\downarrow_s\circledast \mathrm{L}_\zeta(k)||^2_F$. We use the Adam optimizer for 60 epochs with $\mathrm{lr}=10^{-4}$ and weight decay set to $10^{-4}$.

\subsection{$A^+$ calculation}
\label{sec:amas_calculation}
To use the affine projection proposed in Section~\ref{sec:sr}, we implement the $A^+$ operator using a CNN with two convolutional layers with $32$ filters followed by leaky ReLU ($\alpha=0.2$) activation and a final sub-pixel convolution. The sizes of the kernels are $7\times7$, $5\times5$, and $3\times3$, respectively. Following the approach in \cite{sonderby2016amortised}, we train this network by minimizing the following loss function with a stochastic gradient descent, that is,
\begin{align}
 \hat \mu&=\hspace*{-0.1cm}\argmin\mu \E_{x}\|Ax - AA_\mu^+(Ax)\|_2^2 \nonumber\\
 &+ \E_{y}\|A_\mu^+(y) - A_\mu^+(AA_\mu^+(y))\|_2^2,
\end{align}
where $A$ is the degradation operator and $A_\mu^+$ is a CNN of the parameters $\mu$. We stop the optimization when the loss value is less than $10^{-7}$. During the training of our main network, we keep $\mu$ fixed.

\begin{table*}[t]
\begin{center}
\setlength{\tabcolsep}{2pt}
\begin{tabular}{|c|c|c|c|c|c|c|c|c|c|}
    \hline
    1 & 2 & 3 & 4 & 5 & 6 & 7 & 8 & 9 & 10\\\hline
    Model & $\mathrm{P}_{\psi}(y)$ & $\mathrm{P}_{\psi}(y, \mathrm{PCA}(k))$ & $\mathrm{P}_{\psi}(y_{w})$ &
    $\mathrm{P}_{\psi}\mathrm{D}_{\phi}(y_{w})$-no-PI & $\mathrm{P}_{\psi}\mathrm{D}_{\phi}(y_{w})$ &
    $\mathrm{F}_{\omega}\mathrm{U}_{\psi}\mathrm{D}_{\phi}(y_{w})$ & $\mathrm{F}_{\omega}\mathrm{U}_{\psi}\mathrm{D}_{\phi}(y, y_{w})$ & CNN-Cascade & CADUF \\\hline
    PSNR/SSIM & 26.63-0.773 & 27.06-0.780 & 27.09-0.785 & 27.24-0.789 & 27.37-0.792 & 27.51-0.795 & \underline{27.53-0.796} & 27.25-0.788 & {\bf 27.90-0.806} \\\hline
\end{tabular}
\end{center}
\caption{Key component analysis of the proposed model for scale factor 4 on the GaussianSM testing dataset (images from BSD100 \cite{bsd100}, Urban100 \cite{huang2015self} and Manga109 \cite{Matsui2017} datasets). Best and second-best results appear in bold and underlined, respectively.}
\label{tab:ablation}
\end{table*}

\subsection{Datasets}
\label{sec:data}
The training dataset is formed by images from the DIV2K \cite{DIV2K} and Flickr2K \cite{Flickr2K} datasets. The ground-truth of our training consists of 3450 high-quality 2K images. During training, random patches cropped from the images of the dataset were extracted. The size of each HR patch is $s48\times s48$, where $s$ is the scaling factor. The LR images were synthesized according to Eq.~(\ref{eq:im_for}). For testing, we used images from standard SR test image datasets BSD100 \cite{bsd100}, Urban100 \cite{huang2015self}, and Manga109 \cite{Matsui2017}. To ensure that a wide variety of degradations are present in the test, we generated three LR images for each of the HR images.

The blur kernels used to generate the LR images cause a uniform blur and are a combination of an isotropic Gaussian blur kernel and a motion blur kernel. We simulate two different motion blur kernels representing simple and complex camera movements. In the simple case, only simple linear movement is considered. Meanwhile, the motion blur kernels for the complex case simulate more complex camera movements with curved trajectories. They are generated using the method in \cite{Boracchi:2012} with $T=0.8$ and anxiety $10^R/1000$, where $R$ is a random number from a uniform $[0,1]$ distribution. Furthermore, in this scenario, AWGN of standard deviation $\sigma = 0.01$ is added to the LR image. The combination of Gaussian blur and the motion blur kernels of each scenario produces two sets of blur kernels that we use in our experiments. Fig.\ref{fig:example_kernels} in Columns (1,3,6) and (2,4,5,7) show blur kernel examples of the simple and complex case respectively. We refer to these two sets of kernels, GaussianSM and GaussianCM, respectively.

In the case of GaussianSM, the $\sigma$ of the Gaussian blur is sampled randomly during training from the range [0.2, 2.0] and [0.2, 4.0] for scaling factors 2 and 4, respectively. The motion blur kernels of GaussianSM are sampled with angles in the $[0, 180)$ range and length in the range $[1, 9]$ and $[1, 15]$ for scaling factors 2 and 4, respectively. GaussianCM consists of 640 blur kernels with sizes between $11\times 11$ and $45 \times 45$ pixels
that are combined with a Gaussian blur with $\sigma$ in the range [0.2, 1.0] and [0.2, 2.0] for factors 2 and 4, respectively. We used 540 of these kernels for training and 100 for testing.

\subsection{Model training}
\label{sec:training}
We train our models using the Adam optimizer with $\beta_1 =0.9$, $\beta_2=0.999$, weight decay of $10^{-4}$, and a batch size of 64. For each epoch, we sample 3000 batches. We define {\cal L} as the Charbonnier loss \cite{MSLapSRN}, as we found it to be more stable than MSE. The Charbonnier loss between two images $u$ and $v$ is $\gamma(u, v) = \sum_i \sum_j \sqrt{(u_{i,j} -v_{i,j})^2 + \varepsilon ^2}$, where $\varepsilon=10^{-3}$. We apply data augmentation by randomly performing horizontal and vertical flips, 90-degree rotations, and random scaling with a factor between [0.5, 1.25]. The parameter of the Wiener filter is set as the minimum between $0.001$ and the estimated standard deviation of the noise in the images. All of our models use the RGB images as input and do not require any additional preprocessing step. The implementation was performed using Pytorch \cite{pytorch2019}\footnote{The code and trained models shown in this work will be made available upon acceptance of the paper at https://github.com/vipgugr/CADUF.}.

Our training method consists of two distinctive phases:
A) An initialization phase where we train our model using much higher $\alpha=0.6$ and $\beta=0.3$, forcing the model to first converge to good solutions for the intermediary problems of an optimal LR image and initial high-res image estimation. Although it is possible to train the proposed model without this initial step, we have found that this initialization helps stabilize the model and makes it converge earlier. We trained for 20 epochs with $\mathrm{lr}=10^{-4}$.

B) Main training phase. We set $\alpha=0.1$ and $\beta=0.1$ and train our model for 120 epochs. The learning rate was set to $10^{-4}$ for the first 90 epochs and then set to $10^{-5}$ for the remaining 30 epochs.\\

\subsection{Ablation study}
\label{sec:ablation}
In this section, we conducted experiments to determine the contribution of each component of the proposed CADUF. All models have 28 residual blocks and are trained as described in Section~\ref{sec:training}. The experiments were conducted for scale factor 4 and using GaussianSM training and testing sets (see Section~\ref{sec:data}),with the difference that the blurs used were sampled with $\sigma$ in the range of [0.2, 3.0) and length of $[1, 11]$. This was done to ease the training on the simpler models because using strong blurs makes it unstable. Table~\ref{tab:ablation} contains the results for this study in terms of the PSNR and SSIM.

The base model $\mathrm{P}_{\psi}(y)$ is a single CNN model that uses the architecture described in Section~\ref{sec:architecture_detail} (see Fig~\ref{fig:architecture_detail}), where $h$ is the features extracted from $y$ with two convolutional layers and $z=y$. This model shows the performance that can be expected from a classic SR CNN model. We improve the model performance by progressively adding each main component of the CADUF.

The first improvement of 0.46dB-0.12(SSIM) is achieved by introducing the $k^L$ blur information into the network using the Weiner-filtered image $y_w$ instead of $y$ ($\mathrm{P}_{\psi}(y_w)$) (Columns 2 and 4 in Table.\ref{tab:ablation}). Facilitating the blur information eases the network task because the blur no longer needs to be identified. Alternatively, Zhang et al. \cite{zhang2018learning} proposed using PCA to encode the information of the blur kernel $k$ and use it alongside the blur image $y$. We also test this approximation, making the model $\mathrm{P}_{\psi}(y, \mathrm{PCA}(k))$ (Column 3 in Table.\ref{tab:ablation}). Despite the noise and artifacts introduced by the Weiner filter and the approximation error of using Eq.~(\ref{degradation_new}), $\mathrm{P}_{\psi}(y_w)$ slightly outperforms $\mathrm{P}_{\psi}(y, \mathrm{PCA}(k))$. Moreover, the use of PCA requires its training with the possible blurs $k$, which harms the generalization capabilities of the model to unseen blurs.

The second improvement is the decomposition of $\mathrm{P}_{\psi}(y_w)$ into $\mathrm{P}_{\psi}\mathrm{D}_{\phi}(y_w)$ (Columns 4 and 6 in Table.\ref{tab:ablation}). By specializing part of the model in the denoising and deconvolution of the LR image, we can see an increase in performance of 0.28dB-0.007(SSIM) despite the use of the same number of residual blocks (a total of 28 residual blocks, eight residual blocks for $\mathrm{D}_{\phi}$ and 20 for $\mathrm{P}$). We also tested the contribution of using privileged information (PI) \cite{vapnik2015privi} during training. By training $\mathrm{D}_{\phi}$ to map the input to $x\downarrow_s$ instead of $x\downarrow_s^\mathrm{D}$, we obtain a new model that does not make use of PI. We call this model $\mathrm{P}_{\psi}\mathrm{D}_{\phi}(y_w)$-no-PI. As seen in Columns 5 and 6 of Table~\ref{tab:ablation}, the use of PI is a key element of the proposed CADUF because omitting it significantly deteriorates the performance (0.13dB-0.003(SSIM)).

Next, we use the approximation proposed in Section~\ref{sec:sr} to separate $\mathrm{P}_{\psi}$ into $\mathrm{F}_{\omega}\mathrm{U}_{\psi}$, obtaining the model $\mathrm{F}_{\omega}\mathrm{U}_{\psi}\mathrm{D}_{\phi}(y_w)$ (Column 7-Table~\ref{tab:ablation}). Using the same number of residual blocks (28), $\mathrm{F}_{\omega}\mathrm{U}_{\psi}\mathrm{D}_{\phi}(y_w)$ outperforms $\mathrm{P}_{\psi}\mathrm{D}_{\phi}(y_w)$ (Column 6-Table~\ref{tab:ablation}) by 0.14dB-0.003(SSIM). This is in line with our hypothesis that the decomposition of the task of $\mathrm{P}_{\psi}$ into two sub-tasks, upsampling, and error correction, simplifies the optimization and allows for better minima. It can also be observed that, owing to the introduction of prior information and the use of constraints, the concatenation of $\mathrm{D}_{\phi}$ and $\mathrm{U}_{\psi}$ does not output complex artifacts that are difficult to filter out by $\mathrm{F}_{\omega}$. As shown in \cite{ulyanov18deep}, CNN architectures introduce deep priors, which in some applications explain a large portion of the model efficiency. However, in our case, the ablation study shows that it is the specific training, with the introduction of $\mathrm{F}_{\omega}\mathrm{U}_{\psi}$ and privileged information, that is the primary factor responsible for the model efficiency.

We also tested the proposed mechanism to connect all the sub-modules of the CADUF. To do that, we develop a new model by feeding each sub-module only with the image $z$ calculated by the previous one (not the features $h$) and omitting the reuse of $z$ with dynamic filtering. By doing this, we obtain a classic CNN-cascade model similar to those used in the literature \cite{Yuan2018UnsupervisedIS,gu2019blind}, a concatenation of a deblur and upsampling CNN. This model, which we call CNN-Cascade (Column 9 in Table.\ref{tab:ablation}), obtains results far worse than $\mathrm{F}_{\omega}\mathrm{U}_{\psi}\mathrm{D}_{\phi}(y_w)$ and even $\mathrm{P}_{\psi}\mathrm{D}_{\phi}(y_w)$ (0.26dB-0.007(SSIM) and 0.12dB-0.004(SSIM), respectively). As we argue in Section~\ref{sec:model_description}, the difference in performance is because the use of the features of previous sub-modules allows for better information propagation through the cascade.

Finally, we evaluate the contribution of our proposed motion-corrected feature extraction sub-module, $\mathrm{E}_ {\theta}$, appending the information extracted from $y$ and $y_w$ together. Overall, it constitutes the full CADUF model (Column 10 in Table.\ref{tab:ablation}). Compared to the concatenation of $y$ and $y_w$ ($\mathrm{F}_{\omega}\mathrm{U}_{\psi}\mathrm{D}_{\phi}(y, y_w)$) (Column 8 in Table.\ref{tab:ablation}), the use of $\mathrm{E}_ {\theta}$ significantly increases the performance (0.37dB-0.010(SSIM)) with little overhead computation.

\subsection{SOTA Comparison}
\label{sec:SOTA}

\begin{table*}[t]
\begin{center}
\begin{tabular}{|c|c| c | c | c | c | c | c | c |}
    \hline
    \multirow{3}{*}{Scale} & \multirow{3}{*}{Model} & \multicolumn{6}{c|}{PSNR-SSIM} & \multirow{3}{*}{MACs} \\\cline{3-8}
    & & \multicolumn{3}{c|}{GaussianSM} & \multicolumn{3}{c|}{GaussianCM} & \\\cline{3-8}
    & & BSD100 \cite{bsd100} & Urban100 \cite{huang2015self} & Manga109 \cite{Matsui2017} & BSD100 \cite{bsd100} & Urban100 \cite{huang2015self} & Manga109 \cite{Matsui2017} & \\\hline\hline
    \multirow{9}{*}{$\times2$} & Bicubic & 26.21-0.692 & 23.58-0.685 & 25.54-0.809 & 23.24-0.543 & 20.68-0.529 & 21.28-0.655 & $2.02*10^{7}$ \\\cline{2-9}
    & ZSSR \cite{ZSSR} & 26.55-0.717 & 23.81-0.711 & 25.76-0.810 & 24.11-0.556 & 21.79/0.591 & 21.92-0.664 & $4.70*10^{13}$ \\\cline{2-9}
    & RCAN \cite{Zhang_etcol_2018}* & 30.85-0.874 & 29.22-0.882 & 34.27-0.956 & 25.13-0.640 & 22.61-0.639 & 24.40-0.775 & $2.46*10^{12}$ \\\cline{2-9}
    & IRCNN \cite{Zhang2017learning}+RCAN \cite{Zhang_etcol_2018} & 27.08-0.759 & 24.66-0.751 & 27.92-0.879 & 26.23-0.712 & 24.30-0.732 & 27.01-0.846 & $2.70*10^{12}$ \\\cline{2-9}
    & DeblurGAN+RCAN \cite{Zhang_etcol_2018} & 26.21-0.702 & 23.57-0.694 & 25.54-0.815 & 23.99-0.573 & 21.55-0.568 & 22.86-0.706 & $2.47*10^{12}$ \\\cline{2-9}
    & SRMDNF \cite{zhang2018learning}* & 30.83-0.873 & 29.42-0.884 & 34.94-0.957 & 25.99-0.695 & 24.01-0.712 & 26.36-0.825 & $\mathbf{2.60*10^{11}}$ \\\cline{2-9}
    & SFTMD \cite{gu2019blind}* & \underline{31.57-0.889} & \underline{30.62-0.907} & \underline{36.79-0.969} & \underline{27.26-0.742} & 25.66-0.774 &  28.97-0.880 & $9.71*10^{11}$ \\\cline{2-9}
    & DPSR \cite{zhang2019deep} & 28.37-0.793 & 27.11-0.815 & 33.19-0.925 & 27.23-0.740 & \underline{25.89-0.790} & \underline{30.11-0.895} & $4.86*10^{12}$ \\\cline{2-9}
    & CADUF & {\bf 31.92-0.893} & {\bf 31.11-0.914} & {\bf 37.70-0.972} & {\bf 27.92-0.761} & {\bf 26.64-0.805} & {\bf 30.73-0.906} & $\underline{6.56*10^{11}}$ \\\hline\hline
    \multirow{9}{*}{$\times4$} & Bicubic & 24.19-0.582 & 21.51-0.564 & 22.52-0.698 & 23.05-0.533 & 20.40-0.511 & 20.99-0.642 & $1.68*10^{7}$ \\\cline{2-9}
    & ZSSR \cite{ZSSR} & 24.76-0.589 & 21.83-0.573 & 22.88-0.709 & 23.32-0.545 & 20.90-0.520 & 21.45-0.652 & $2.14*10^{13}$ \\\cline{2-9}
    & RCAN \cite{Zhang_etcol_2018}* & 26.92-0.713 & 24.86-0.736 & 28.80-0.877 & 24.80-0.610 & 22.35-0.612 & 24.13-0.756 & $6.38*10^{11}$ \\\cline{2-9}
    & IRCNN \cite{Zhang2017learning}+RCAN \cite{Zhang_etcol_2018} & 24.71-0.596 & 22.18-0.587 & 23.60-0.714 & 24.19-0.567 & 21.87-0.565 & 23.21-0.691 & $6.98*10^{12}$ \\\cline{2-9}
    & DeblurGAN+RCAN \cite{Zhang_etcol_2018} & 23.48-0.550 & 21.34-0.538 & 22.31-0.663 & 22.11-0.498 & 20.59-0.494 & 21.35-0.618 & $6.40*10^{11}$ \\\cline{2-9}
    & SRMDNF \cite{zhang2018learning}* & 26.82-0.708 & 24.76-0.729 & 28.53-0.870 & 25.31-0.632 & 23.10-0.649 & 25.45-0.793 & $\mathbf{6.71*10^{10}}$ \\\cline{2-9}
    & SFTMD \cite{gu2019blind}* & \underline{27.12-0.721} & \underline{25.25-0.752} & \underline{29.42-0.889} & \underline{25.67-0.647} & \underline{23.62-0.674} & \underline{26.46-0.821} & $2.82*10^{11}$ \\\cline{2-9}
    & DPSR \cite{zhang2019deep} & 25.06-0.641 & 23.47-0.686 & 27.35-0.842 & 24.79-0.629 & 22.51-0.656 & 24.95-0.804 & $1.33*10^{12}$ \\\cline{2-9}
    & CADUF & {\bf 27.41-0.729} & {\bf 25.64-0.768} & {\bf 30.05-0.900} & {\bf 25.73-0.652} & {\bf 23.82-0.689} & {\bf 26.78-0.835} & $\underline{2.03*10^{11}}$ \\\hline
\end{tabular}
\end{center}
\caption{Comparison with Non-blind SOTA on the BSD100 \cite{bsd100}, Urban100 \cite{huang2015self} and Manga109 \cite{Matsui2017} datasets for the blur kernels in GaussianSM and GaussianCM datasets. To ensure fairness in the comparison, the methods indicated with ``*'' were retrained. The average number of MACs per image of each method is also reported. Best and second-best results are bold and underlined, respectively.}
\label{tab:sota_motion_non_blind}
\end{table*}

In this section, we compare our CADUF model against the SOTA methods in SISR dealing with multiple degradations. RCAN \cite{Zhang_etcol_2018} is a very deep CNN developed for SISR that does not use any information about the degradation but achieves a very high score. In contrast, SRMDNF \cite{zhang2018learning} and SFTMD \cite{gu2019blind} incorporate information of the kernel $k$ into the network by encoding it using PCA, but obtaining lower performance than RCAN \cite{Zhang_etcol_2018}. We also show the results of combining SOTA deconvolution methods with an SISR CNN in a cascade. Specifically, we compared two combinations, both using deconvolution followed by upsampling: IRCNN \cite{Zhang2017learning}+RCAN \cite{Zhang_etcol_2018} and DeblurGAN+RCAN \cite{Zhang_etcol_2018}. Finally, we also compare the MBO approaches ZSSR \cite{ZSSR} and DPSR \cite{zhang2019deep}.

\begin{figure*}[!htb]
\captionsetup[subfigure]{labelformat=empty}
\begin{center}
\setlength{\tabcolsep}{1pt} 
\renewcommand{\arraystretch}{0}
\resizebox{0.98\textwidth}{!}{
\begin{tabular}{ccccc}
\multirow{4}{*}[3.4em]{\subfloat[ ``HighschoolKimengumi\_vol01'']{\includegraphics[width=0.28\textwidth]{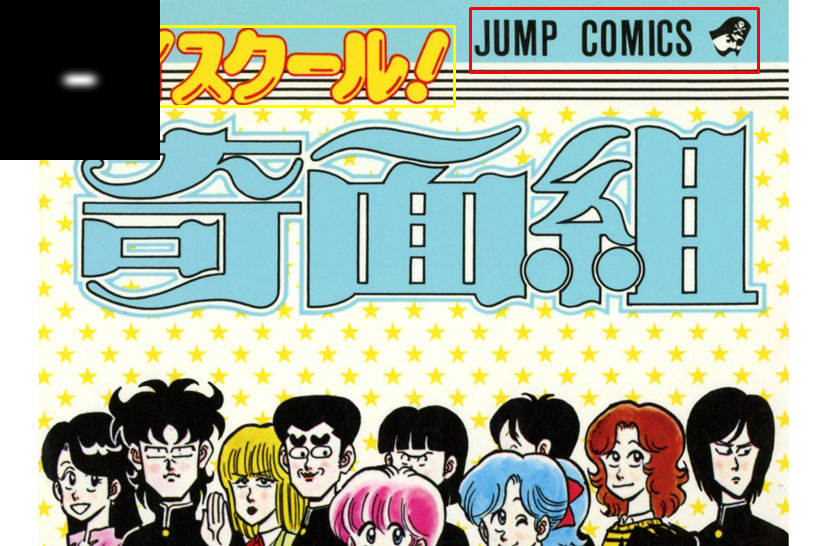}}} &

\subfloat[Bicubic]{
\begin{tabular}{c}
\includegraphics[width=0.18\textwidth]{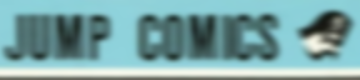} \\
\includegraphics[width=0.18\textwidth]{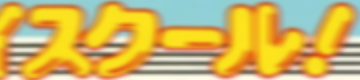}
\end{tabular}} &

\subfloat[IRCNN \cite{Zhang2017learning}+RCAN \cite{Zhang_etcol_2018}]{
\begin{tabular}{c}
\includegraphics[width=0.18\textwidth]{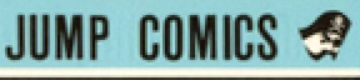} \\
\includegraphics[width=0.18\textwidth]{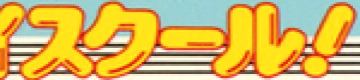}
\end{tabular}} &

\subfloat[RCAN \cite{Zhang_etcol_2018}]{
\begin{tabular}{c}
\includegraphics[width=0.18\textwidth]{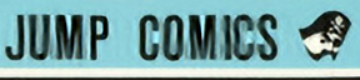} \\
\includegraphics[width=0.18\textwidth]{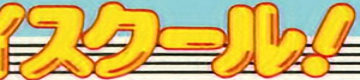}
\end{tabular}} &

\subfloat[SRMDNF \cite{zhang2018learning}]{
\begin{tabular}{c}
\includegraphics[width=0.18\textwidth]{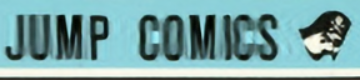} \\
\includegraphics[width=0.18\textwidth]{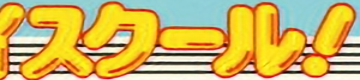}
\end{tabular}} \\

&\subfloat[SFTMD \cite{gu2019blind}]{
\begin{tabular}{c}
\includegraphics[width=0.18\textwidth]{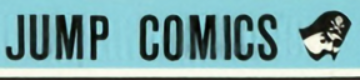} \\
\includegraphics[width=0.18\textwidth]{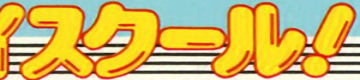}
\end{tabular}} &

\subfloat[DPSR \cite{zhang2019deep}]{
\begin{tabular}{c}
\includegraphics[width=0.18\textwidth]{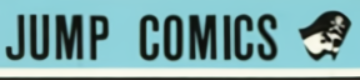} \\
\includegraphics[width=0.18\textwidth]{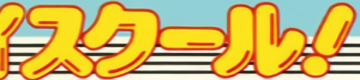}
\end{tabular}} &

\subfloat[CADUF]{
\begin{tabular}{c}
\includegraphics[width=0.18\textwidth]{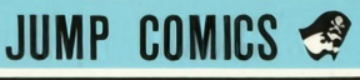} \\
\includegraphics[width=0.18\textwidth]{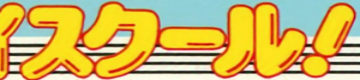}
\end{tabular}} &

\subfloat[HR]{
\begin{tabular}{c}
\includegraphics[width=0.18\textwidth]{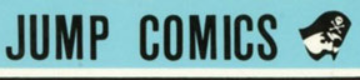} \\
\includegraphics[width=0.18\textwidth]{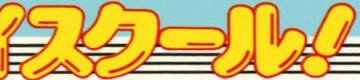}
\end{tabular}} \\ [3.4em]

\multirow{4}{*}[3.55em]{\subfloat[ ``img\_049'']{\includegraphics[width=0.28\textwidth]{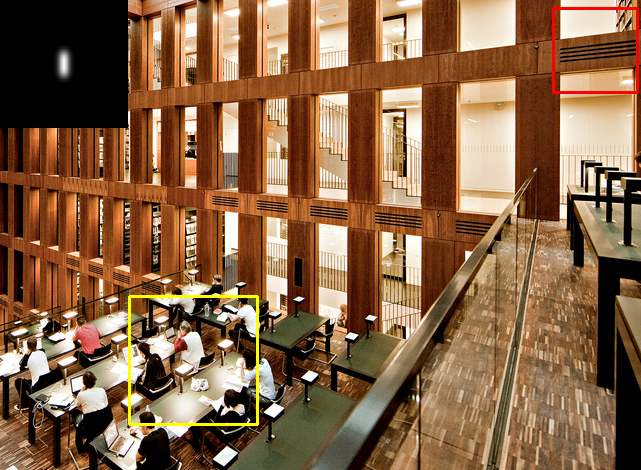}}} &

\subfloat[Bicubic]{
\begin{tabular}{cc}
\includegraphics[width=0.09\textwidth]{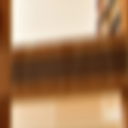} &
\includegraphics[width=0.09\textwidth]{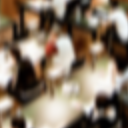}
\end{tabular}} &

\subfloat[IRCNN \cite{Zhang2017learning}+RCAN \cite{Zhang_etcol_2018}]{
\begin{tabular}{cc}
\includegraphics[width=0.09\textwidth]{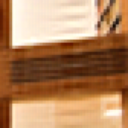} &
\includegraphics[width=0.09\textwidth]{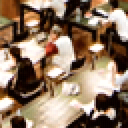}
\end{tabular}} &

\subfloat[RCAN \cite{Zhang_etcol_2018}]{
\begin{tabular}{cc}
\includegraphics[width=0.09\textwidth]{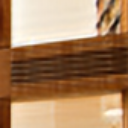} &
\includegraphics[width=0.09\textwidth]{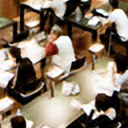}
\end{tabular}} &

\subfloat[SRMDNF \cite{zhang2018learning}]{
\begin{tabular}{cc}
\includegraphics[width=0.09\textwidth]{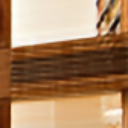} &
\includegraphics[width=0.09\textwidth]{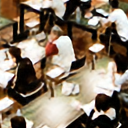}
\end{tabular}} \\

&\subfloat[SFTMD \cite{gu2019blind}]{
\begin{tabular}{cc}
\includegraphics[width=0.09\textwidth]{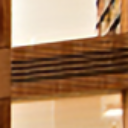} &
\includegraphics[width=0.09\textwidth]{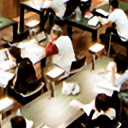}
\end{tabular}} &

\subfloat[DPSR \cite{zhang2019deep}]{
\begin{tabular}{cc}
\includegraphics[width=0.09\textwidth]{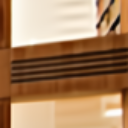} &
\includegraphics[width=0.09\textwidth]{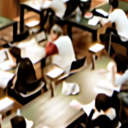}
\end{tabular}} &

\subfloat[CADUF]{
\begin{tabular}{cc}
\includegraphics[width=0.09\textwidth]{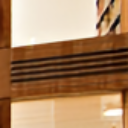} &
\includegraphics[width=0.09\textwidth]{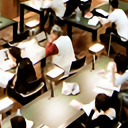}
\end{tabular}} &

\subfloat[HR]{
\begin{tabular}{cc}
\includegraphics[width=0.09\textwidth]{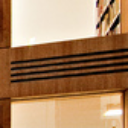} &
\includegraphics[width=0.09\textwidth]{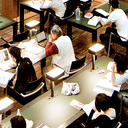}
\end{tabular}} \\ [3.4em]

\multirow{4}{*}[3.55em]{\subfloat[ ``58060'']{\includegraphics[width=0.28\textwidth]{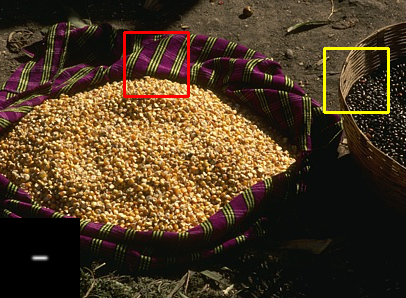}}} &

\subfloat[Bicubic]{
\begin{tabular}{cc}
\includegraphics[width=0.09\textwidth]{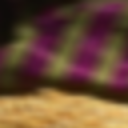} &
\includegraphics[width=0.09\textwidth]{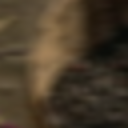}
\end{tabular}} &

\subfloat[IRCNN \cite{Zhang2017learning}+RCAN \cite{Zhang_etcol_2018}]{
\begin{tabular}{cc}
\includegraphics[width=0.09\textwidth]{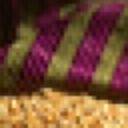} &
\includegraphics[width=0.09\textwidth]{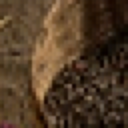}
\end{tabular}} &

\subfloat[RCAN \cite{Zhang_etcol_2018}]{
\begin{tabular}{cc}
\includegraphics[width=0.09\textwidth]{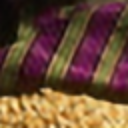} &
\includegraphics[width=0.09\textwidth]{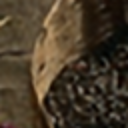}
\end{tabular}} &

\subfloat[SRMDNF \cite{zhang2018learning}]{
\begin{tabular}{cc}
\includegraphics[width=0.09\textwidth]{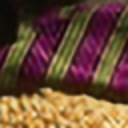} &
\includegraphics[width=0.09\textwidth]{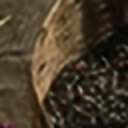}
\end{tabular}} \\

&\subfloat[SFTMD \cite{gu2019blind}]{
\begin{tabular}{cc}
\includegraphics[width=0.09\textwidth]{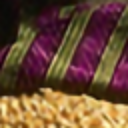} &
\includegraphics[width=0.09\textwidth]{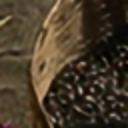}
\end{tabular}} &

\subfloat[DPSR \cite{zhang2019deep}]{
\begin{tabular}{cc}
\includegraphics[width=0.09\textwidth]{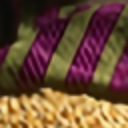} &
\includegraphics[width=0.09\textwidth]{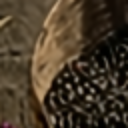}
\end{tabular}} &

\subfloat[CADUF]{
\begin{tabular}{cc}
\includegraphics[width=0.09\textwidth]{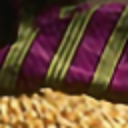} &
\includegraphics[width=0.09\textwidth]{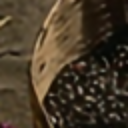}
\end{tabular}} &

\subfloat[HR]{
\begin{tabular}{cc}
\includegraphics[width=0.09\textwidth]{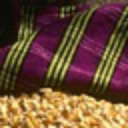} &
\includegraphics[width=0.09\textwidth]{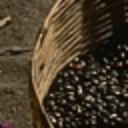}
\end{tabular}} \\

\end{tabular}}
\end{center}
\caption{Comparison of IRCNN \cite{Zhang2017learning}+RCAN \cite{Zhang_etcol_2018}, RCAN \cite{Zhang_etcol_2018}, SRMDNF \cite{zhang2018learning}, SFTMD \cite{gu2019blind}, DPSR \cite{zhang2019deep} and CADUF on GaussianSM dataset for factor 2 on image ``HighschoolKimengumi\_vol01'' from Managa109 \cite{Matsui2017} dataset, image ``img\_049'' from Urban100 \cite{huang2015self} dataset and image ``58060'' from BSD100 \cite{bsd100} dataset.}
\label{fig:image_results_sm}
\end{figure*}

\begin{figure*}[!htb]
\captionsetup[subfigure]{labelformat=empty}
\begin{center}
\setlength{\tabcolsep}{1pt}
\renewcommand{\arraystretch}{0}
\resizebox{0.98\textwidth}{!}{
\begin{tabular}{cccc}
\subfloat[``PsychoStaff'' HR]{
\includegraphics[width=0.25\textwidth]{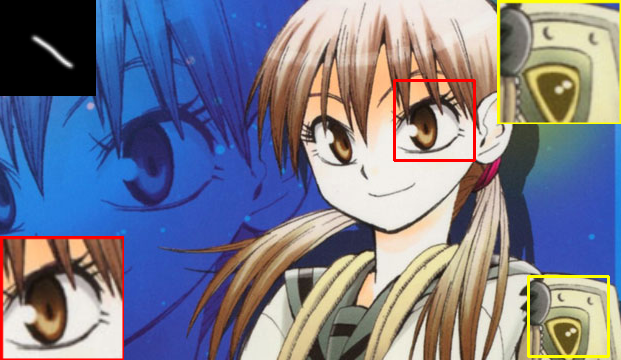}} &
\subfloat[Bicubic]{
\includegraphics[width=0.25\textwidth]{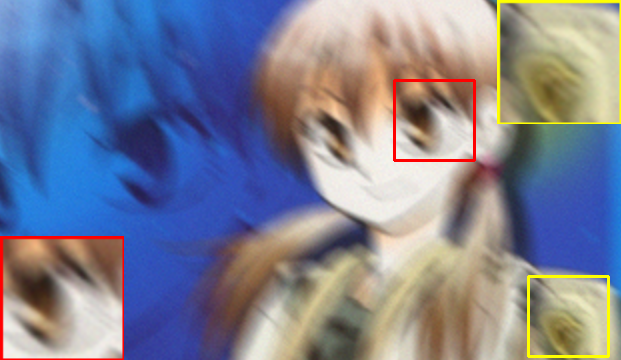}} &
\subfloat[IRCNN \cite{Zhang2017learning}+RCAN \cite{Zhang_etcol_2018}]{
\includegraphics[width=0.25\textwidth]{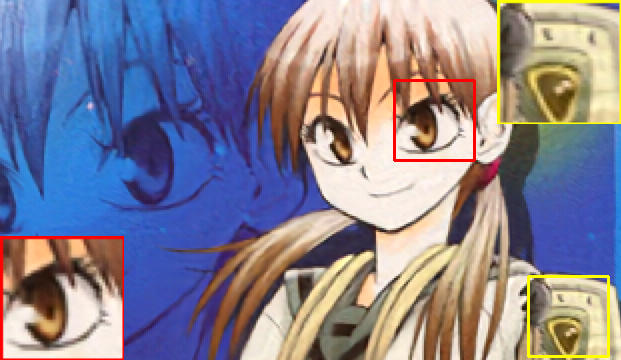}} &
\subfloat[RCAN \cite{Zhang_etcol_2018}]{
\includegraphics[width=0.25\textwidth]{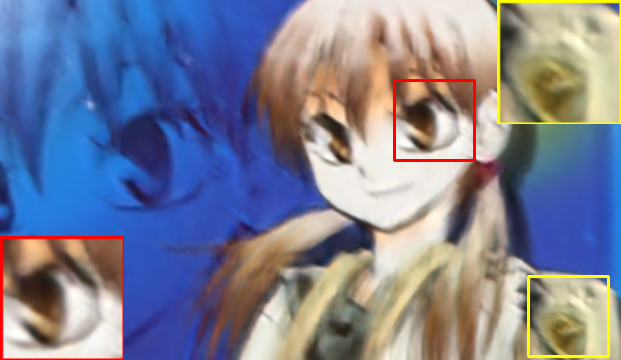}} \\
\subfloat[SRMDNF \cite{zhang2018learning}]{
\includegraphics[width=0.25\textwidth]{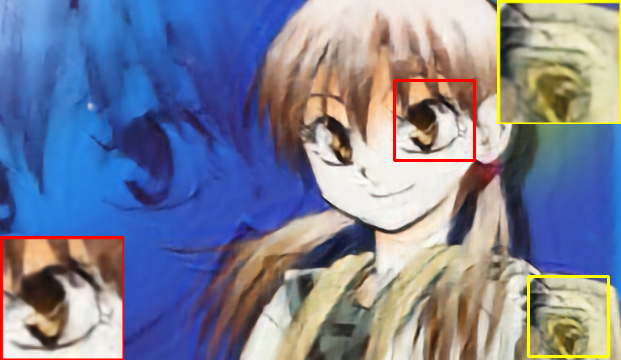}} &
\subfloat[SFTMD \cite{gu2019blind}]{
\includegraphics[width=0.25\textwidth]{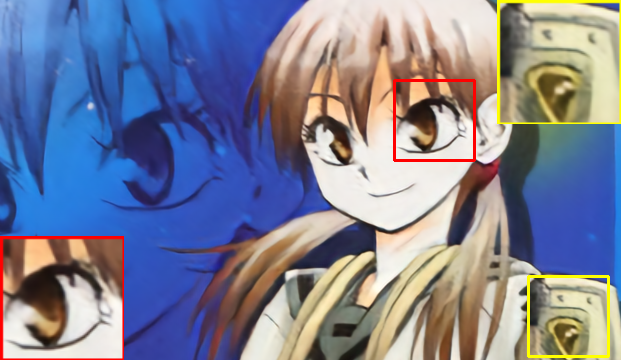}} &
\subfloat[DPSR \cite{zhang2019deep}]{
\includegraphics[width=0.25\textwidth]{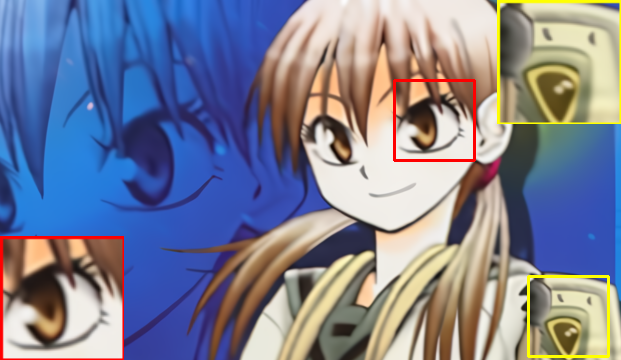}} &
\subfloat[CADUF]{
\includegraphics[width=0.25\textwidth]{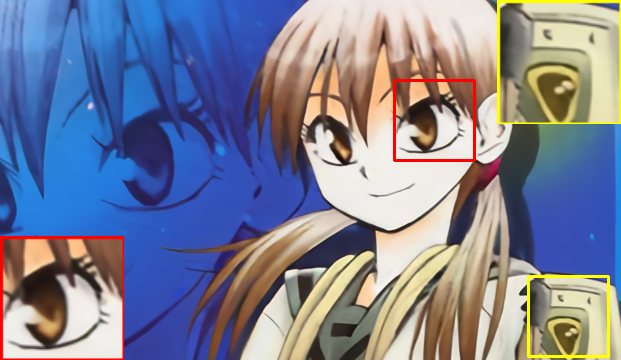}} \\
\subfloat[``img\_089'' HR]{
\includegraphics[width=0.25\textwidth]{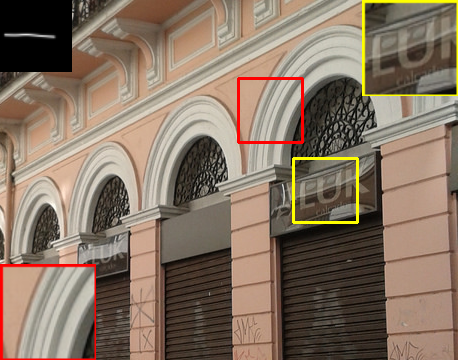}} &
\subfloat[Bicubic]{
\includegraphics[width=0.25\textwidth]{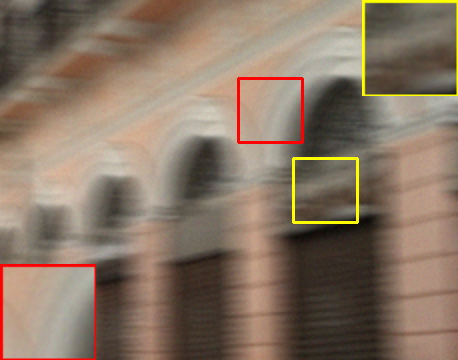}} &
\subfloat[IRCNN \cite{Zhang2017learning}+RCAN \cite{Zhang_etcol_2018}]{
\includegraphics[width=0.25\textwidth]{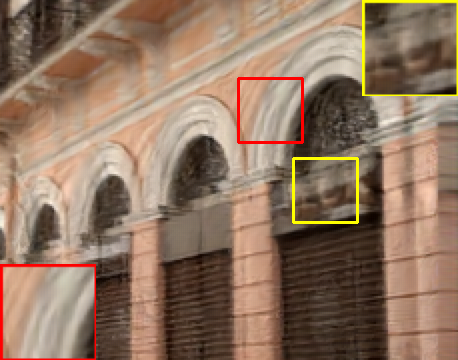}} &
\subfloat[RCAN \cite{Zhang_etcol_2018}]{
\includegraphics[width=0.25\textwidth]{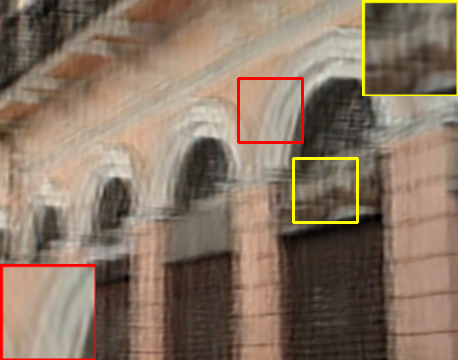}} \\
\subfloat[SRMDNF \cite{zhang2018learning}]{
\includegraphics[width=0.25\textwidth]{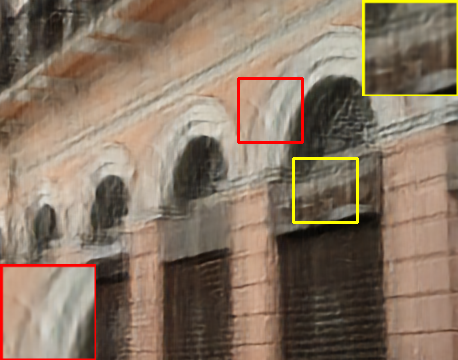}} &
\subfloat[SFTMD \cite{gu2019blind}]{
\includegraphics[width=0.25\textwidth]{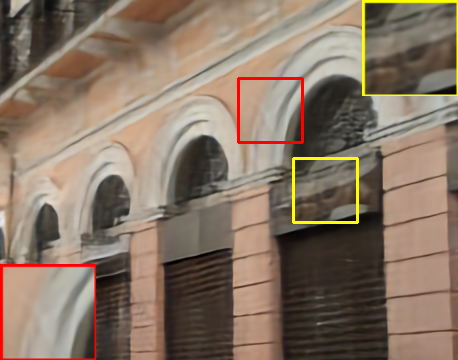}} &
\subfloat[DPSR \cite{zhang2019deep}]{
\includegraphics[width=0.25\textwidth]{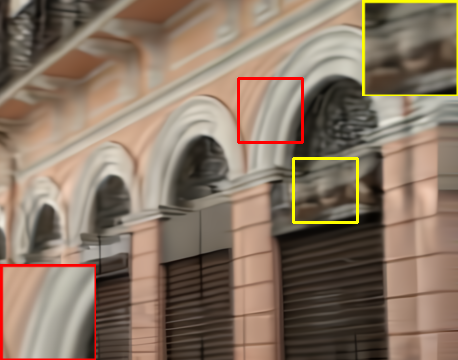}} &
\subfloat[CADUF]{
\includegraphics[width=0.25\textwidth]{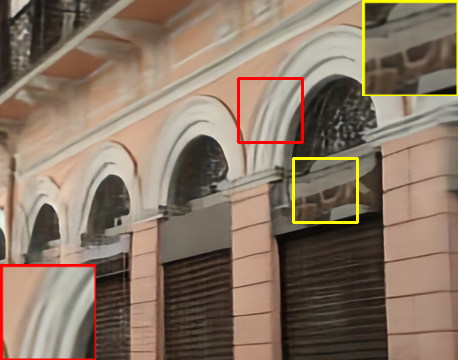}} \\
\subfloat[``img\_096'' HR]{
\includegraphics[width=0.25\textwidth]{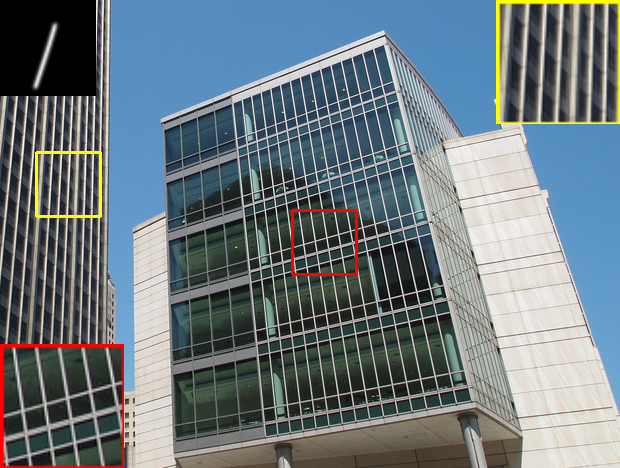}} &
\subfloat[Bicubic]{
\includegraphics[width=0.25\textwidth]{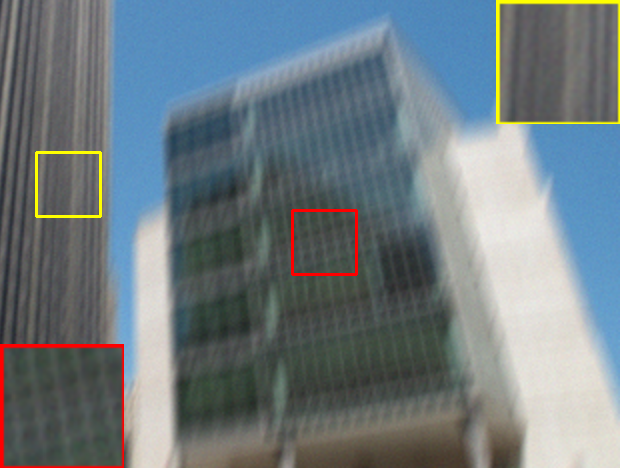}} &
\subfloat[IRCNN \cite{Zhang2017learning}+RCAN \cite{Zhang_etcol_2018}]{
\includegraphics[width=0.25\textwidth]{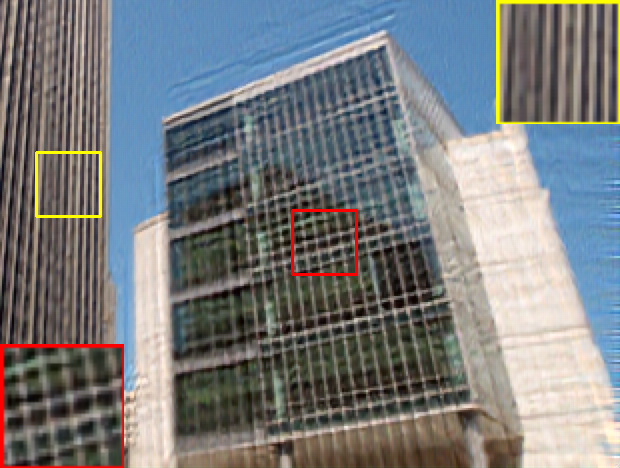}} &
\subfloat[RCAN \cite{Zhang_etcol_2018}]{
\includegraphics[width=0.25\textwidth]{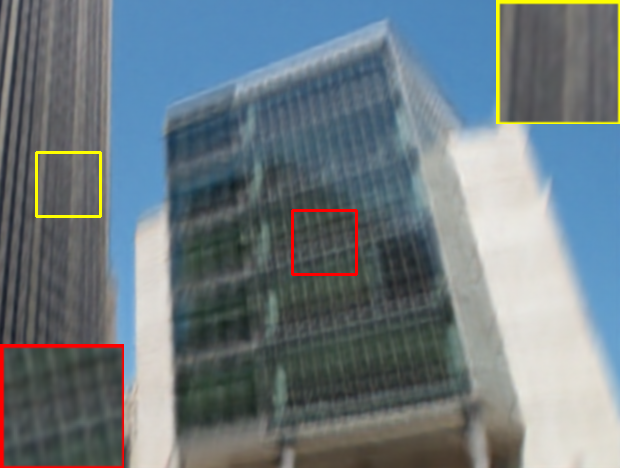}} \\
\subfloat[SRMDNF \cite{zhang2018learning}]{
\includegraphics[width=0.25\textwidth]{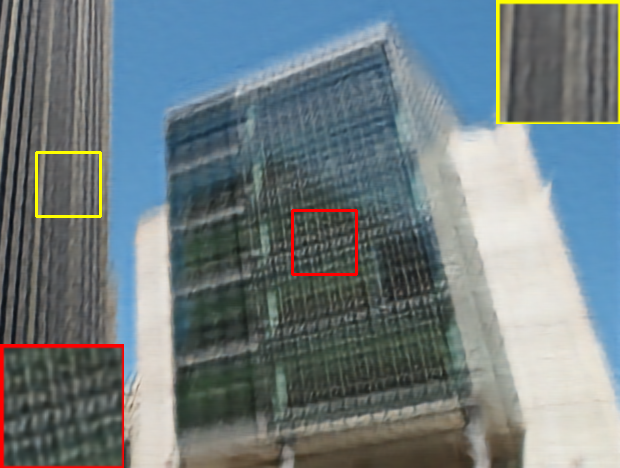}} &
\subfloat[SFTMD \cite{gu2019blind}]{
\includegraphics[width=0.25\textwidth]{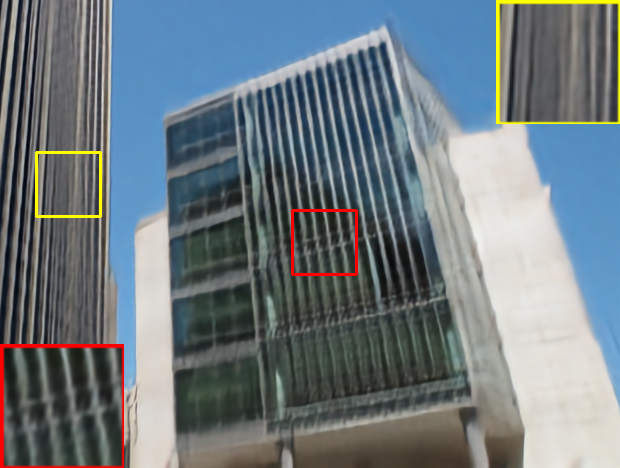}} &
\subfloat[DPSR \cite{zhang2019deep}]{
\includegraphics[width=0.25\textwidth]{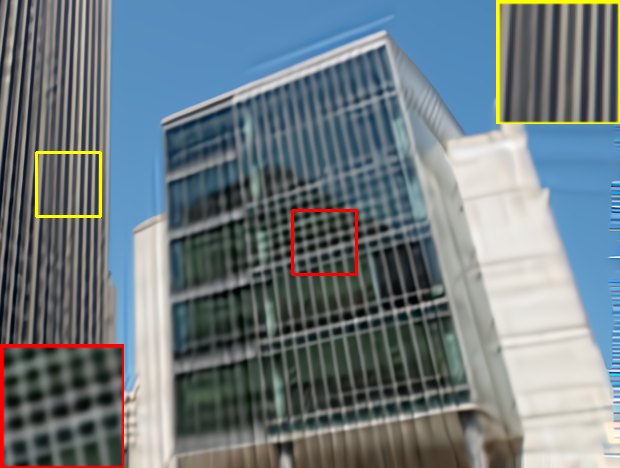}} &
\subfloat[CADUF]{
\includegraphics[width=0.25\textwidth]{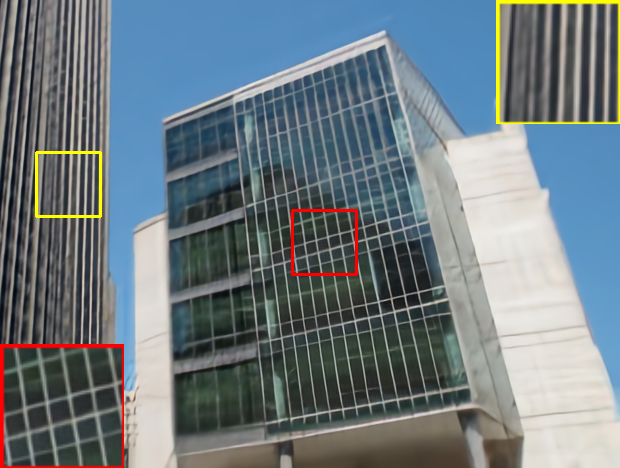}} \\
\end{tabular}}
\end{center}
\caption{Visual comparison of IRCNN \cite{Zhang2017learning}+RCAN \cite{Zhang_etcol_2018}, RCAN \cite{Zhang_etcol_2018}, SRMDNF \cite{zhang2018learning}, SFTMD \cite{gu2019blind}, DPSR \cite{zhang2019deep} and CADUF on GaussianCM dataset for factor 2 on image ``PsychoStaff'' from Managa109 \cite{Matsui2017} dataset and images ``img\_089'' and ``img\_096'' from Urban100 \cite{huang2015self} dataset.}
\label{fig:image_results_cm}
\end{figure*}

We test all models using the test datasets GaussianSM and GaussianCM described in Section~\ref{sec:data}. To ensure fairness in the comparison, we not only train our model for each setting using the appropriate training set, but also do the same for competing models that require it. Therefore, we trained RCAN \cite{Zhang_etcol_2018}, SRMDNF \cite{zhang2018learning}, and SFTMD \cite{gu2019blind}. The training and evaluation of the competing methods was performed using the code provided by the original authors.

Table~\ref{tab:sota_motion_non_blind} shows the results of our CADUF model and competing models in the non-blind scenario for GaussianSM and GaussianCM kernels datasets and scaling factors 2 (x2) and 4 (x4). We evaluate the performance of the methods using the PSNR and SSIM metrics. Furthermore, we also include a study of the time complexity of each method using the average number of multiplication and additions (MACs) per image as a metric. The smaller the number of MACs, the faster the algorithm performance.

As seen in Table~\ref{tab:sota_motion_non_blind}, compared to other LB methods, CADUF performs significantly better in all datasets and scenarios. In GaussianSM, CADUF obtains an average improvement on the datasets of 2.13dB-0.02(SSIM) in x2 and 0.84dB-0.02(SSIM) in x4 over RCAN \cite{Zhang_etcol_2018} and 0.58dB-0.01(SSIM) in x2 and 0.44dB-0.01(SSIM) in x4 over SFTMD \cite{gu2019blind}, being significantly faster. The performance difference increases for the GaussianCM dataset. In this case, the figures show improvements of 4.38dB-0.14(SSIM) in x2 and 1.68dB-0.06(SSIM) in x4 over RCAN \cite{Zhang_etcol_2018} and 1.14dB-0.03(SSIM) in x2 and 0.19dB-0.01(SSIM) in x4 over SFTMD \cite{gu2019blind}.

\begin{table}[t]
\begin{center}
\setlength{\tabcolsep}{2pt}
\begin{tabular}{|c|c|c|c|c|}
    \hline
    \multirow{2}{*}{Scale} & \multirow{2}{*}{Model} & \multicolumn{3}{c|}{PSNR-SSIM} \\\cline{3-5}
    & & BSD100 \cite{bsd100} & Urban100 \cite{huang2015self} & Manga109 \cite{Matsui2017} \\\hline\hline
    \multirow{6}{*}{$\times2$}
    & Bicubic & 23.24-0.543 & 20.68-0.529 & 21.28-0.655 \\\cline{2-5}
    & RCAN \cite{Zhang_etcol_2018}* & \underline{25.13}-0.640 & \underline{22.61}-0.639 & \underline{24.40}-0.775 \\\cline{2-5}
    & DeblurGAN+RCAN \cite{Zhang_etcol_2018} & 23.99-0.573 & 21.55-0.568 & 22.86-0.706 \\\cline{2-5}
    & IKC \cite{gu2019blind}+SFTMD \cite{gu2019blind}* & 18.27-0.451 & 16.29-0.456 & 17.50/0.585 \\\cline{2-5}
    & DPSR \cite{zhang2019deep} & 23.56-\underline{0.678} & 20.32-\underline{0.652} & 23.57-\underline{0.825} \\\cline{2-5}
    & Our Model & {\bf 25.65-0.698} & {\bf 23.27-0.674} & {\bf 25.06-0.847} \\\hline
\end{tabular}
\end{center}
\caption{Comparison with blind SOTA on the BSD100 \cite{bsd100}, Urban100 \cite{huang2015self} and Manga109 \cite{Matsui2017} datasets for the blur kernels in the GaussianCM dataset. To ensure fairness in the comparison, the methods indicated with ``*'' were retrained. Best and second-best results are bold and underlined, respectively.}
\label{tab:sota_motion_blind}
\end{table}

The difference between CADUF and the other best performing methods in GaussianSM is more apparent if we perform a closer inspection of the produced images in Fig.~\ref{fig:image_results_sm}. As shown, when uniform motion blur is present in the image, CADUF is able to recover fine details in the image as well as the other best performing method, SFTMD \cite{gu2019blind}, but without introducing ghosting and other artifacts. Despite being a blind method, RCAN \cite{zhang2018learning} is able to compete with the remaining methods in GaussianSM, obtaining similar results to other non-blind methods such as SRMDNF \cite{zhang2018learning}. However, its performance drops significantly for the more complex blurs in GaussianCM, showing that even a CNN as complex as RCAN \cite{zhang2018learning} is not able to cope with the degradation variance of a more complex realistic scenario without the introduction of specialized elements. In this case, although SRMDNF \cite{zhang2018learning} and SFTMD \cite{gu2019blind} do not suffer as much as RCAN \cite{zhang2018learning}, they show a more significant drop in performance than CADUF and DPSR \cite{zhang2019deep}. We believe that this is because both rely on PCA to encode blur information. Since the variability of the blur is much higher, it is more difficult to accurately depict new blur kernels at test time. Furthermore, both models rely on standard convolutional layers that are not well suited for recovering blurs caused by strong camera motions. In contrast, the use of the motion-corrected feature extraction sub-module and the Wiener filter allows our model to not only adapt to blur introduced by more complex camera motions but also to better generalize and be more robust to blurs not seen in training. These features are not shared by the other two cascade models IRCNN \cite{Zhang2017learning}+RCAN \cite{Zhang_etcol_2018} and DeblurGAN+RCAN \cite{Zhang_etcol_2018}. Both produce worse results than RCAN \cite{Zhang_etcol_2018} and cannot compete with the remaining CNN methods in terms of performance and time consumption. The accumulated errors in the cascade and the strong artifacts that appear due to it are the source of this poor performance.

\begin{figure}[!htb]
\captionsetup[subfigure]{labelformat=empty}
\begin{center}
\setlength{\tabcolsep}{1pt} 
\renewcommand{\arraystretch}{0}
\resizebox{0.48\textwidth}{!}{
\begin{tabular}{cccc}
\subfloat[``376043'' HR]{
\includegraphics[width=0.125\textwidth]{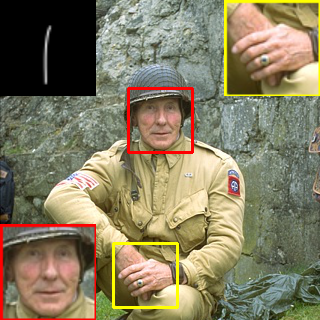}} &
\subfloat[Bicubic]{
\includegraphics[width=0.125\textwidth]{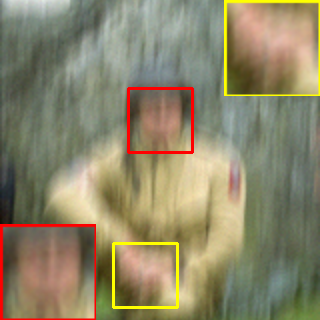}} &
\subfloat[IRCNN \cite{Zhang2017learning}+RCAN \cite{Zhang_etcol_2018}]{
\includegraphics[width=0.125\textwidth]{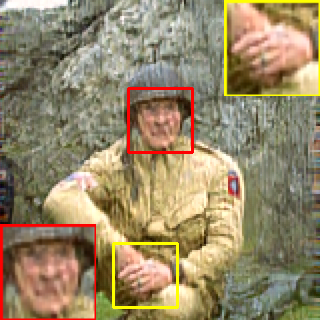}} &
\subfloat[RCAN \cite{Zhang_etcol_2018}]{
\includegraphics[width=0.125\textwidth]{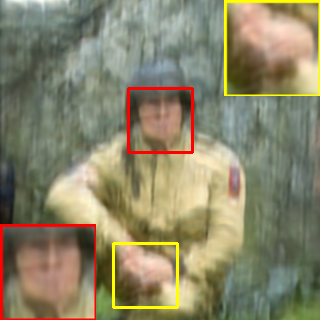}} \\
\subfloat[SRMDNF \cite{zhang2018learning}]{
\includegraphics[width=0.125\textwidth]{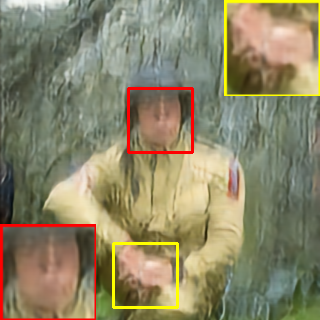}} &
\subfloat[SFTMD \cite{gu2019blind}]{
\includegraphics[width=0.125\textwidth]{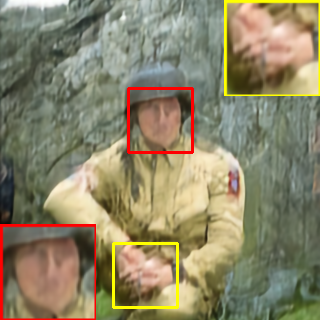}} &
\subfloat[DPSR \cite{zhang2019deep}]{
\includegraphics[width=0.125\textwidth]{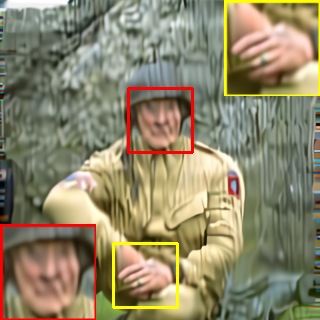}} &
\subfloat[CADUF]{
\includegraphics[width=0.125\textwidth]{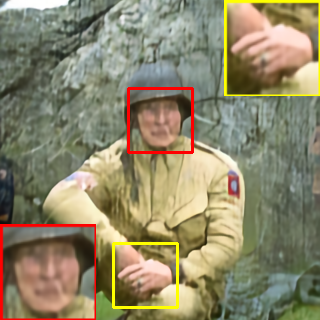}} \\
\end{tabular}}
\end{center}
\caption{Failure case of CADUF for factor 2 on image ``376043'' from the BSD100 \cite{bsd100} dataset with degradation from the GaussianCM dataset.}
\label{fig:image_results_fail}
\end{figure}

Compared to MBO methods (ZSSR \cite{ZSSR} and DPSR \cite{zhang2019deep}), we can see that CADUF requires orders of magnitude fewer MACs because it does not need to solve an expensive optimization problem for each new test image. Although our model's ability to resolve unseen blurs is lower, it generalizes well to new degradations close to those observed in training, that is, a similar type of noise and blur (motion, defocus, downsampling). We can see that our model significantly outperforms both ZSSR \cite{ZSSR} and DPSR \cite{zhang2019deep}, with the latter being the closest in terms of performance of the other two. Fig.~\ref{fig:image_results_cm}\footnote{More examples can be downloaded at https://github.com/vipgugr/CADUF.} shows a visual comparison of SOTA methods and CADUF for testing images of the GaussianCM dataset. As seen, compared to the other two best-performing models, SFTMD \cite{gu2019blind} and DPSR \cite{zhang2019deep}, CADUF produces significantly better images, closer to the HR image and with far fewer artifacts. Especially remarkable is the case of image ``img096'' of the Urban100 \cite{huang2015self} dataset, where CADUF is able to reconstruct the building windows on the center of the image far better than the other methods. Despite the good performance shown by CADUF, when very strong blurs are present, it would fail to properly recover the HR image. One example of these cases is shown in Fig.~\ref{fig:image_results_fail}. The CADUF method is unable to recover the face of the soldier and produces some easily seen artifacts in the background wall and the soldier helmet. However, CADUF still produces better results than the other SOTA methods, introducing significantly fewer artifacts and better recovering the hands of the soldier and more details of the face. When making a comparative visual study of our model against the SOTA, we found that it produces competing results when the blur does not include too much motion and outperforms other models when strong motion is present.

Finally, in Table~\ref{tab:sota_motion_blind}, we compare our method against competing methods in the blind setting in the GaussianCM dataset. We estimate the LR blur kernel $k^{*L}$ for algorithms that require it (DPSR \cite{zhang2019deep} and ours) using the method in \cite{Zhou2017}. In our model case, we fine-tuned it for 100 epochs with lr=$10^{-5}$. We report results for factor 2 because the loss of information coupled with errors during the degradation estimation makes it difficult to analyze the differences between the methods for factor 4. As seen, the proposed model has a significantly less drop in performance compared to DPSR \cite{zhang2019deep} when an approximation of $k^{*L}$ is used. In this case, our model benefits of the possibility to fine-tune and adapt to the errors in the blur kernel estimation. Thanks to this fact, it still significantly outperforms all other methods (2.18dB-0.02(SSIM) and 0.61dB-0.06(SSIM) over DPSR \cite{zhang2019deep} and RCAN \cite{Zhang_etcol_2018}, respectively).

\section{Conclusions}
\label{sec:conclusions_results}
In this study, a new cascade CNN model for SISR capable of dealing with multiple degradations, has been proposed. The model exhibits a new approach for searching for a solution with several differentiating keys. First, cascade sub-modules are explicitly specialized in specific tasks: motion-corrected feature extraction, deblur, and upsampling. Second, a new way of regularizing the deblur and upsampling sub-modules is introduced from domain knowledge. As a consequence of the upsampling restriction using affine projection \cite{sonderby2016amortised}, a final sub-module that corrects the accumulated cascade error can be introduced to further increase the model performance. As far we know, the use of domain knowledge to constrain the training of modules in a cascade represents a novelty in SISR. Finally, the model uses an improved method of connecting the sub-modules in the cascade where each sub-module re-uses the features calculated by all previous sub-modules. As a result, our model is fast, robust, and accurate.
The experiments were performed using two new SISR datasets, containing degradations with simple and complex camera movement. These experiments show that CADUF significantly outperforms CNN-BMs with architectures of the same depth. Furthermore, CADUF outperforms the remaining compared SOTA methods for non-blind and blind SISR while being significantly faster. Although CADUF is a CNN-BM, it can be generalized to unseen blurs during training, competing in this regard with MBO models.

\ifCLASSOPTIONcaptionsoff
  \newpage
\fi

\bibliographystyle{ieeetr}
\bibliography{ms}

\end{document}